\documentclass{article}

\usepackage[preprint]{neurips_2020}

\usepackage[utf8]{inputenc} 
\usepackage[T1]{fontenc}    
\usepackage{hyperref}       
\usepackage{url}            
\usepackage{booktabs}       
\usepackage{amsfonts}       
\usepackage{nicefrac}       
\usepackage{microtype}      
\usepackage{amssymb}
\usepackage{amsmath}
\usepackage{algorithm}
\usepackage{algorithmic}
\usepackage{graphicx,subcaption}
\usepackage{xcolor}
\usepackage{tabularx}
\usepackage{rotating}

\title{Reinforcement Learning for Ridesharing: An Extended Survey\thanks{This survey is a significantly expanded and refined version of \citep{qin2021reinforcement}. Please cite as Zhiwei (Tony) Qin, Hongtu Zhu, and Jieping Ye. Reinforcement learning for ridesharing: An extended survey. Transportation Research Part C: Emerging Technologies,
Volume 144,
2022.}}

\author{%
  Zhiwei (Tony) Qin\thanks{Corresponding author, email: zq2107@caa.columbia.edu. The original version of this paper was completed while the author was at DiDi Labs, Mountain View, CA, USA. Any views presented in this paper do not represent those of either DiDi or Lyft.} \\
  Lyft Rideshare Labs \\
  San Francisco, CA 94107 \\
   \And
   Hongtu Zhu \\
   University of North Carolina \\
   Chapel Hill, NC 27514 \\
   \AND
   Jieping Ye \\
   University of Michigan \\
   Ann Arbor, MI 48109 \\
}

\begin{document}

\maketitle

\begin{abstract}
In this paper, we present a comprehensive, in-depth survey of the literature on reinforcement learning approaches to decision optimization problems in a typical ridesharing system. Papers on the topics of rideshare matching, vehicle repositioning, ride-pooling, routing, and dynamic pricing are covered. Most of the literature has appeared in the last few years, and several core challenges are to continue to be tackled: model complexity, agent coordination, and joint optimization of multiple levers.
Hence, we also introduce popular data sets and open simulation environments to facilitate further research and development. Subsequently, we discuss a number of challenges and opportunities for reinforcement learning research on this important domain.
\end{abstract}

\section{Introduction}
The emergence of ridesharing\footnote{The business model covered in this survey is also often referred to as `ride-sourcing' or `on-demand ride services' \citep{wang2019ridesourcing}.}, led by companies such as DiDi, Uber, and Lyft, has revolutionized the form of personal mobility. It is projected that the global rideshare industry will grow to a total market value of \$218 billion by 2025 \citep{markets2018ridesharevol}. 
However, how to improve operational efficiency is a major challenge for rideshare platforms, e.g., long passenger waiting time \citep{wtop2019wait} and as high as 41\% vacant time for ridesharing vehicles in a large city \citep{wsj2020vacant}. The success of ridesharing, from the perspectives of the platforms, drivers, and passengers, requires sophisticated optimization of all the integrated components that collectively deliver the services.

Reinforcement learning (RL) is a machine learning paradigm that trains an agent to take optimal actions (measured by total cumulative reward) through interaction with the environment and getting feedback signals. It is a class of optimization methods for solving sequential decision-making problems with a long-term objective in a stochastic environment. Thanks to the rapid advancement in deep learning research and computing power, the integration of deep neural networks and RL has generated explosive progress in solving complex large-scale decision problems \citep{silver2016alphago,berner2019dota}, attracting huge amount of renewed interests in the recent years. We are witnessing a similar trend in the ridesharing domain, where the demand and supply are highly stochastic and non-stationary, and the operational decisions are often sequential in nature and have strong spatiotemporal dependency. Simple greedy heuristics that only optimizes for immediate returns tend to produce short-sighted reactive policies, which do not align well with cumulative nature of the true performance measure of interest. The multi-step sequential nature of the decision-making (e.g., pricing, matching, repositioning) and the supply-demand stochasticity in the environment pose enormous challenges to traditional predictive and optimization methods, spanning such issues as forecast accuracy, decision-time computational complexity, and adaptability to real-time changes.
RL, on the other hand, presents itself as an excellent promising approach to these ridesharing  optimization problems. RL methods are often highly data-driven, making them more suitable to situations where it is hard to build accurate predictive models. They are forward-looking, and yet, they do not explicitly depend on forecasting. And, by design, RL-based policies are dynamic and often light in decision-time complexity.

There are excellent surveys on RL for intelligent transportation \citep{haydari2020survey,yau2017tscsurvey}, with in-depth coverage of traffic signals control and autonomous driving. \citep{wang2019ridesourcing} offers a broad review of ridesharing systems, whereas \cite{tong2020spatial} surveys spatial crowdsourcing, which is a more general field than ridesharing.  There has been no comprehensive review of the literature on RL for ridesharing, even though the field has attracted much attention and interest from the research communities for both RL and transportation just within the last few years (e.g., \citep{shah2020neural,tang2019deep,al2020approximate,shou2020reward}). This paper aims to fill that gap by surveying the literature of this domain published in top conferences and journals in transportation (e.g., Transportation Research series, IEEE Transactions on Intelligent Transportation Systems, Transportation Science), data mining (e.g., KDD, ICDM, WWW, CIKM), and machine learning/AI (e.g., NeurIPS, AAAI, IJCAI). We describe the research problems associated with the various aspects of a ridesharing system, review the existing RL approaches proposed to tackle each of them, and discuss the challenges and opportunities. 
Reinforcement learning also has close relationship with approximate dynamic programming. Although it is not our goal in this paper to have a comprehensive review of this class of  methods for problems in ridesharing, we aim to point the readers to the representative works so that they can refer to the further literature therein.

This survey is organized as follows: We lay out the ridesharing system architecture in Section \ref{sec:rideshare} and define the scope of the problems to be reviewed. Within this section, as well as the subsequent sections of the survey, we clarify and draw connections among the different names that the problems are referred to, which are often due to the different communities that the authors are from and are easy to confuse by researchers new to ridesharing. In Section \ref{sec:rl}, we provide a concise review of the RL basics and the major algorithms adopted by the works in this survey. We review in details in Section \ref{sec:survey} the literature for each problem described in Section \ref{sec:rideshare} and the relevant data sets and environments. Finally in Section \ref{sec:discussions}, we discuss some challenges and opportunities that we feel crucial in advancing RL for ridesharing.

\section{Ridesharing}\label{sec:rideshare}
We first describe the architecture of a ridesharing system in this section, followed by explanation and clarification on the scopes of the problem associated with each module.

\subsection{Architecture}\label{sec:architecture}
A ridesharing service, in contrast to taxi hailing, matches passengers with drivers of vehicles for hire using mobile apps. 
In a typical mobile ridesharing system, there are five major modules: pricing, matching, repositioning, pooling, and routing. Figure \ref{fig:rideshare_process} illustrates the process and decision modules.
When a potential passenger submits a trip request, the \emph{pricing} module offers a quote, which the passenger either accepts or rejects. Upon acceptance, the \emph{matching} module attempts to assign the request to an available driver. Depending on driver pool availability, the request may have to wait in the system until a successful match. Pre-match cancellation may happen during this time. The assigned driver then travels to pick up the passenger, during which time post-match cancellation may also occur. The pick-up location is usually where the passenger is making the request or he/she specifies. In some cases, it could be a public designated area, e.g., outside an airport or train station. After the driver successfully transports the passenger to the destination, she receives the trip fare and becomes available again. The \emph{repositioning} module guides idle vehicles to specific locations in anticipation of fulfilling more requests in the future. Following the reposition recommendations is usually on a voluntary basis unless it is an autonomous ridesharing setting. Hence, it is common that the  platform offers incentives to drivers for completing the repositions. 
When each driver takes only one passenger request at a time, i.e. only one passenger shares the ride with the driver, this mode is more commonly called `ride-hailing'. Ridesharing can refer to both ride-hailing and ride-pooling.\footnote{In this survey, we do not cover topics on hitch, in which the driver is on his/her own trip with a specific destination.}
In the \emph{ride-pooling} mode, multiple passengers with different trip requests can share one single vehicle, so the pricing, matching, repositioning, and routing problems are different from those for ride-hailing and require specific treatment, in particular, considering the passengers already on board.
The \emph{routing} module provides turn-by-turn guidance on the road network to drivers/vehicles either in service of a passenger request or performing a reposition. The goal is to guide the vehicle to its destination efficiently and safely.  

\subsection{Problem Scopes}\label{sec:scope}
\begin{figure}
\begin{center}
        \includegraphics[width=0.85\linewidth]{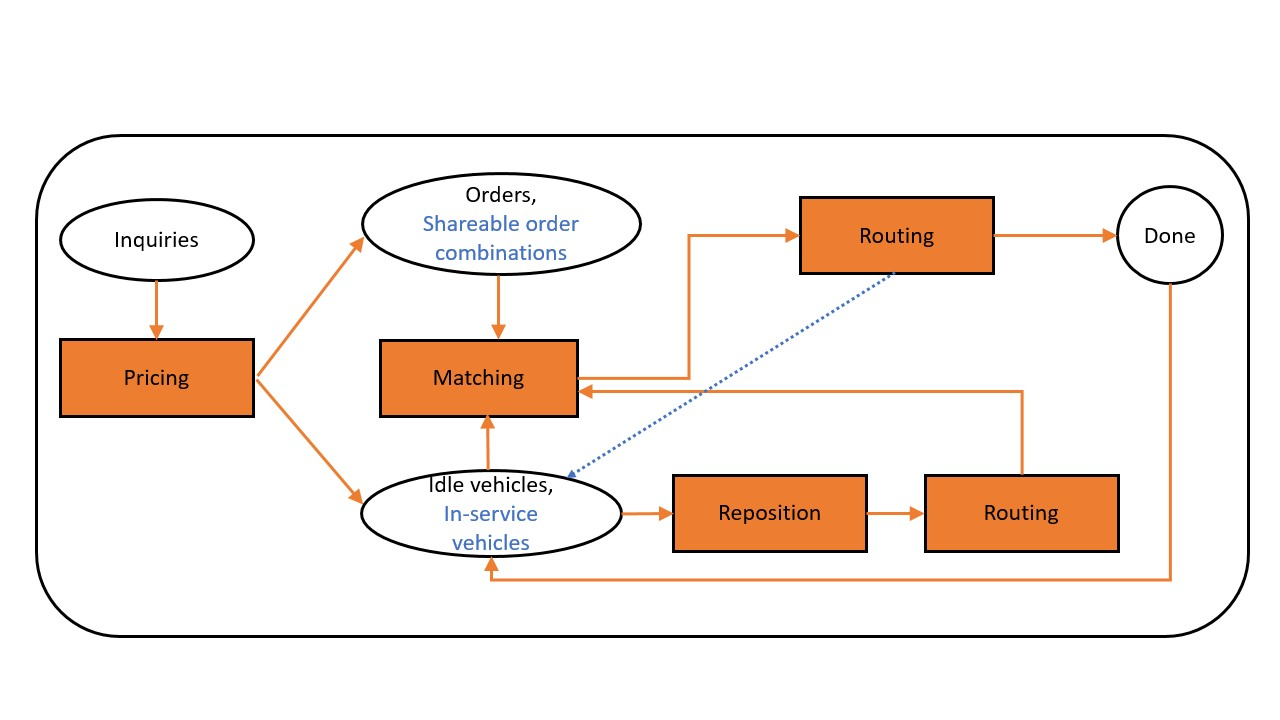}
	\caption{The process flow of ridesharing operations. The solid orange rectangular boxes represent the modules described in Section \ref{sec:rideshare}, and the literature on the optimization problems associated with the modules are reviewed in the paper. The blue text and arrow apply exclusively to ride-pooling to account for the fact that order combinations and in-service vehicles are also eligible to participate in matching.}
	\label{fig:rideshare_process}
\end{center}
\end{figure}

First, we start from the pricing module.
Since the trip fare is both the price that the passenger has to pay for the trip and the major factor for the income of the driver, pricing decisions influence both demand and supply distributions through price sensitivities of users, e.g., the use of surge pricing during peak hours. This is illustrated by the solid arrows pointing from the pricing module to orders and idle vehicles respectively in Figure \ref{fig:rideshare_process}.
The pricing problem in the ridesharing literature is in most cases dynamic pricing, which adjusts trip prices in real-time in view of the changing demand and supply. The pricing modules sits at the upstream position with respect to the other modules and is a macro-level lever to achieve supply-demand (SD) balance. Although technically, driver pay can be determined by a separate module from pricing and has its own implication on supply elasticity and driver behavior, this paper follows the common setting where driver pay is closely associated (approximately proportional) to the trip fare so that pricing has the dual effect on demand and supply.

The ridesharing matching problem \citep{yan2020dynamic,ozkan2017dynamic,qin2020ride} may appear under different names in the literature, e.g., order dispatching \citep{qin2020ride}, trip-vehicle assignment \citep{bei2018algorithms}, and on-demand taxi dispatching \citep{tong2020spatial}. 
It is an online bipartite matching problem where both supply and demand are dynamic, with the uncertainty coming from demand arrivals, travel times, and the entrance-exit behavior of the drivers. Matching can be done continuously in a streaming manner or at fixed review windows (i.e., batching). Sophisticated matching algorithms often leverage demand prediction in some form beyond the actual requests, e.g., the value function in RL.
Online request matching is not entirely unique to ridesharing. Indeed, ridesharing matching falls into the family of more general dynamic matching problems for on-demand markets \citep{hu2022dynamic}. A distinctive feature of the ridesharing problem is its spatiotemporal nature. A driver's eligibility to match and serve a trip request depends in part on her spatial proximity to the request. Trip requests generally take different amount of time to finish, and they change the spatial states of the drivers, affecting the supply distribution for future matching. The drivers and passengers generally exhibit asymmetric exit behaviors in that drivers usually stay in the system for an extended period of time, whereas passenger requests are lost after a much shorter waiting period in general. 

Single-vehicle repositioning may refer to as taxi routing or passenger seeking in the literature.
Taxi routing slightly differs in the setting from repositioning a rideshare vehicle in that a taxi typically has to be at a visual distance from the potential passenger to take the request whereas the matching radius of a mobile rideshare request is considerably longer, sometimes more than a mile. System-level vehicle repositioning, also known as driver dispatching, vehicle rebalancing/reallocation, or fleet management, aims to rebalance the global SD distributions by proactively dispatching idle vehicles to different locations. 
Repositioning and matching are similar to each other in that both relocate a vehicle to a different place as a consequence. In theory, one can treat repositioning as matching a vehicle to a virtual trip request, the destination of which is that of the reposition action, so that both matching and repositioning can be solved in a single problem instance. Typically in practice, these two problems are solved separately because they are separate system modules on most ridesharing platforms with different review intervals and objective metrics among other details.

The routing module described in Section \ref{sec:architecture} performs route guidance, which could be \emph{dynamic routing} or \emph{route planning} depending on the decision points. Dynamic routing is also called  \emph{dynamic route choice}, and route planning is alternatively referred to as the traffic assignment problem \citep{shou2020multi}. 
In some cases, the reposition policy directly provides link-level turn-by-turn guidance with the goal of maximizing the driver's income, thus covering the role of dynamic routing albeit with a different objective. 
Dynamic routing is generally different from the \emph{vehicle routing problem} (VRP) \citep{dantzig1959truck}. In VRP, the set of destinations that the vehicle has to visit is known in advance, and hence, it is a static problem. It mainly concerns with the sequence in which the destinations should be visited, considering the estimated travel time from point to point. In contrast, dynamic routing is associated with a road network, and the decision to make is which outgoing road (link) to follow at each intersection (node). The decision is adaptive to the changing traffic condition on the road network in real time. In the context of ride-pooling, there is another emerging problem in which the dynamic routing decisions with passenger(s) on board have to align with the overall objective of ride-pooling.


Some literature refers to the mode of multiple passengers sharing a ride as `ridesharing'. In this paper, we use term `\emph{ride-pooling}' (or `carpool') to disambiguate the concept, as `ridesharing' can refer to both single- and multiple-passenger rides. The seminal paper of \cite{alonso2017demand} shows that through ride-pooling optimization, the number of taxis required to meet the same trip demand can be significantly reduced with limited impact on passenger waiting times. In a pooling-enabled rideshare system, the matching, repositioning, and pricing modules all have to adapt to additional complexity.
Compared to the regular ride-hailing problem, the one with ride-pooling has considerably more complexity due to the more dynamic nature of the problem and the additional constraints and multiple objectives that have to be considered. 
In this case, the set of available vehicles are augmented, including both idle vehicles and occupied ones not at capacity. It is non-trivial to determine the set of feasible actions (one or more passengers to be assigned to a vehicle) for matching.
Every time a new passenger is to be added to a non-idle vehicle, the route has to be recalculated using a VRP solver to account for the additional pick-up and drop-off, the travel times for all the passengers already on board are updated, and the vehicle capacity, the waiting time and detour distance constraints are checked.  In-service routing in the context of ride-pooling is discussed in Section \ref{sec:carpool}.



\section{Reinforcement Learning}\label{sec:rl}
We briefly review the RL basics and the major algorithms, especially those used in the works reviewed by this survey. For a complete reference, see, e.g., \citep{sutton2018reinforcement}.


\subsection{Basics}

RL is based on the Markov decision process (MDP) framework, where the agent (the decision-making entity) has a \emph{state} $s$ (e.g., the location of a vehicle) in the state space $\mathcal{S}$ and can perform an \emph{action} $a$ (e.g., to dispatch or idle) in the action space $\mathcal{A}$. The action is determined by the agent's policy, $\pi(s):\mathcal{S}\rightarrow \mathcal{A}$. If the policy is stochastic, then $\pi(a|s)$ gives the probability of selecting $a$ given $s$.
After executing the action, the agent receives an immediate \emph{reward} $R(s,a)$ from the environment, and its state changes according to the transition probabilities $P(\cdot | s,a)$. The process repeats until a terminal state or the end of the horizon is reached, giving a sequence of the triplets $(s_t,a_t,r_t)_{t=0}^{t=T}$, where $t$ is the epoch index, $T$ is the final epoch at the terminal state or the end of the horizon, and $r_t$ is a sample of $R$. The objective of the MDP is to maximize the cumulative reward over the horizon. A key quantity to compute is the value function associated with $\pi$,
\begin{equation*}
    V^\pi(s):=E_\pi\left[\sum_{t=0}^{t=T}\gamma^t r_t \bigg\rvert s_0=s\right],
\end{equation*}
which satisfies the Bellman equation, 
\begin{equation}\label{eq:bellman}
    V^\pi(s_t) = \sum_{a_t}\pi(a_t|s_t)\sum_{s_{t+1},r_t} P(s_{t+1},r_t|s_t,a_t)\bigg( r_t(s_t,a_t) + \gamma V^\pi(s_{t+1}) \bigg).
\end{equation}
Similarly, we have the action-value function,
\begin{equation*}
    Q^\pi(s,a):=E_\pi\left[\sum_{t=0}^{t=T}\gamma^t r_t \bigg\rvert s_0=s,a_0=a\right],
\end{equation*}
which conditions on both $s$ and $a$. The optimal  state- and action-values are denoted by $V^*$ and $Q^*$, evaluated at the optimal policy $\pi^*$. The objective of an MDP is to find an optimal policy $\pi^*$ that maximizes the long-term cumulative discounted reward, i.e., $\pi^* := \arg\max_\pi E_s [V^\pi(s)]$.

\subsection{Algorithms}
Given $P$ and $R$, which specifies the MDP, and $\pi$, we can compute $V^\pi$ by iteratively applying the Bellman equation \eqref{eq:bellman},
\begin{equation}
    V^\pi(s)\leftarrow \sum_a\pi(a|s)\sum_{s^\prime,r}P(s^\prime,r|s,a)\bigg(r+\gamma V^\pi(s^\prime)\bigg).
\end{equation}
This is called \emph{policy evaluation}. Using policy evaluation as a sub-routine, we can again iteratively improve the policy through \emph{policy iterations}, which generate a new policy at each outer iteration by acting greedily with respect to $V^\pi$,
\begin{equation}
    a^* \leftarrow \arg\max_a \sum_{s^\prime,r}P(s^\prime,r|s,a)\bigg(r+\gamma V^\pi(s^\prime)\bigg).
\end{equation}
As a special instance, we can collapse the inner policy evaluation loop to a single iteration and compute $V^*(s), \forall s \in \mathcal{S},$ by the \emph{value iterations}, 
\begin{equation}
    V(s)\leftarrow \max_a \sum_{s^\prime,r}P(s^\prime,r|s,a)\bigg(r+\gamma V(s^\prime)\bigg).
\end{equation}
If we estimate $(P,R)$ from data, we have a basic model-based RL method. A model-free method learns the value function and optimizes the policy directly from data without learning and constructing a model of the environment.
A common example is \emph{temporal-difference (TD) learning} \citep{sutton1988learning}, which iteratively updates $V^\pi$ by TD-errors using $\pi$-generated trajectory samples and bootstrapping, 
\begin{equation}\label{eq:td-learning}
    V^\pi(s) \leftarrow V^\pi(s) + \alpha\bigg(r + \gamma V^\pi(s^\prime) - V^\pi(s)\bigg),
\end{equation}
where $s^\prime$ is the next state in the trajectory after $s$, $\alpha$ is the step size (or learning rate), and the term that it scales is the TD error.
If learning the optimal action-values for control is the goal, we similarly have \emph{Q-learning} \citep{watkins1992q}, which updates the action-value function $Q(s,a)$ to approximate $Q^*(s,a)$ by 
\begin{equation}
    Q(s,a) \leftarrow Q(s,a) + \alpha\bigg(r + \gamma\max_{a^\prime} Q(s^\prime,a^\prime) - Q(s,a)\bigg).
\end{equation}
Q-learning is an \emph{off-policy} algorithm, where the behavior policy, which collects the experience data and typically involves exploration, is different from the target policy that we are trying to learn and in this case, is the optimal policy. The \emph{on-policy} counterpart of Q-learning is SARSA, which basically generalizes TD-learning \eqref{eq:td-learning} to the action-value function associated with the behavior policy $\pi$ (same as the target policy):
\begin{equation}
    Q^\pi(s,a) \leftarrow Q^\pi(s,a) + \alpha\bigg(r + \gamma Q^\pi(s^\prime,a^\prime) - Q^\pi(s,a)\bigg),
\end{equation}
where $a^\prime$ is the action executed by the agent at state $s^\prime$ in the experience data.
The \emph{deep Q-network} (DQN) \citep{mnih2015human} approximates $Q(s,a)$ by a neural network $Q_w$ parametrized by $w$ along with a few heuristic techniques like experience replay \citep{lin1992self} and a target network to improve training stability. These techniques are critical to the successful of DQN in playing Atari games and many other applications, due to the deadly triad issue \citep{sutton2018reinforcement} of reinforcement learning when one tries to combine bootstrapping (i.e., TD-learning and Q-learning), off-policy training (i.e., Q-learning), and function approximations (i.e., neural networks), which may lead to instability and divergence.
The algorithms introduced so far are all value-based methods, which focus on learning the value function, and the policy is derived from the learned value function by, e.g., $\arg\max_a Q(s,a)$. Neural network-based value function approximation is important to ridesharing applications because the state is often high dimensional with the incorporation of SD contextual features. Tabular methods suffer from the curse of dimensionality and are not tractable in this case.

A policy-based method directly learns $\pi$ (which is also called the \emph{actor} and parametrized by $\theta$) by performing stochastic gradient descent. The central step is computing the policy gradient (PG), the gradient of the cumulative reward $J(\theta)$ with respect to the policy parameters $\theta$,
\begin{equation}\label{eq:pg}
    \nabla_\theta J(\theta) = \sum_s \mu(s)\sum_a Q^\pi(s,a)\nabla_\theta\pi(a|s,\theta),
\end{equation}
where $\mu$ is the on-policy distribution under $\pi$. A more common (equivalent) form of the PG \eqref{eq:pg} is
\begin{equation}
    \sum_s \mu(s)\sum_a \pi(a|s,\theta) Q^\pi(s,a)\nabla_\theta\log\pi(a|s,\theta),
\end{equation}
which most of the policy-based methods are based on.
REINFORCE \citep{williams1992simple} is a classical PG method, which uses Monte Carlo (MC) rollout to obtains the sample-based update
\begin{equation}
    \theta_{t+1} \leftarrow \theta_t + \alpha G_t \nabla_\theta\log\pi(a|s,\theta),
\end{equation}
where $G_t$ is an MC approximation of $Q^\pi$. 
As in a value-based method, we can also use function approximation for the action values. The function approximator (e.g., a neural network) $Q_w$ is called the \emph{critic}, and the resulting algorithm is an \emph{actor-critic} (AC) method.
It is well-recognized that the `baseline' version of \eqref{eq:pg}, 
\begin{equation}
    \sum_s \mu(s)\sum_a \pi(a|s,\theta) \bigg(Q^\pi(s,a)-b(s)\bigg)\nabla_\theta\log\pi(a|s,\theta),
\end{equation}
where $b(s)$ is an action-independent baseline,
reduces the variance in the sample gradient and helps speed up learning. Since a natural choice of such a baseline is the state value, the critic often learns the advantage function $Q(s,a)-V(s)$, and the method is called Advantage Actor Critic (A2C). The evaluation of an action is based on how good it can be with respect to the average over all actions, the benefit of which is to reduce the high variance in the actor and to stabilize the model. \cite{mnih2016asynchronous} extend A2C to an asynchronous version (A3C) where independent agents interact with their own copy of the environment and update their model parameters with the master copy asynchronously. This architecture enables much more efficient utilization of the CPU cores through parallel threads and hence accelerates the training. The proximal policy optimization (PPO) \citep{schulman2017proximal} optimizes a clipping surrogate objective with respect to the advantage to promote conservative updates to $\pi$ and is a popular choice of training algorithm for RL problems where policy-based methods are suitable (e.g., with continuous actions).

An MDP can be extended to a Markov game involving multiple agents to form the basis for multi-agent RL (MARL). Many MARL algorithms, e.g., \cite{yang2018mean,lowe2017multi} focus on agent communication and coordination, in view of the intractable action space in MARL. 

\subsection{Approximate Dynamic Programming}
A family of methods closely related to RL is approximate dynamic programming (ADP) \citep{powell2007approximate} for solving stochastic dynamic programs (DP), of which the Bellman equation for MDP is an instance. In ADP methods, unlike that typically seen in RL, a post-decision state $s_t^x$ is often defined to represent the intermediate state to which the current state $s_t$ will transition deterministically given the action $a_t$ before the random factors $\omega_t$ (e.g., demand appearance and cancellation) in the environment realize. With $\omega_t$ fully realized, the state transitions into the next pre-decision state $s_{t+1}$. The value function in an ADP method is defined on the post-decision state and is approximated by a particular functional form. Given the approximated values, the original optimization problem is solved to obtain the decision solution for the current time step. Linear function approximation is popular (e.g., \citep{simao2009approximate,yu2019integrated,al2020approximate}) because the dual variables associated with the solution to the current-stage optimization can be used to update the linear function parameters. Then, the state is advanced to the next pre-decision state, and the iteration continues until convergence. By nature, ADP methods are on-policy methods. Recently, neural network-based value function approximation \citep{shah2020neural} has also been adopted and developed due to their higher level of flexibility. In this case, the value function updates largely follow the DQN scheme.
The ADP methods for ridesharing reviewed in this survey solve system-level stochastic DP problems (e.g., matching and repositioning) and aim to approximate the system value by decomposing it into local or driver-centric values, and the update schemes employed fall into the family of approximate value iterations.

\section{Reinforcement Learning for Ridesharing}\label{sec:survey}
We review the RL literature for ridesharing in this section grouped by the core operational problems described in Section \ref{sec:rideshare}. We first cover pricing, matching, repositioning, and routing in the context of ride-hailing. Then, we will review works on those problems specific to ride-pooling. 

\subsection{Pricing}\label{sec:pricing}
RL-based approaches have been developed for dynamic pricing in one-sided retail markets \citep{raju2003reinforcement,bertsimas2006dynamic}, where pricing changes only the demand pattern per customers' price elasticity. The ridesharing marketplace, however, is more complex due to its two-sided nature and spatiotemmporal dimensions. In this case, pricing is also a lever to change the supply (driver) distribution if price changes are broadcast to the drivers. \cite{chen2021spatial} describe examples of such elasticity functions for both demand and supply for their simulation environment.  

The challenges in dynamic pricing for ridesharing lie in both its exogeneity and endogeneity. Dynamic pricing on trip inquiries changes the subsequent distribution of the submitted requests through passenger price elasticity. The requests distribution, in turn, influences future supply distribution as drivers fulfill those requests. On the other hand, the trip fares influence the demand for ridesharing services at given locations, and these changes will affect the pool of waiting passengers, which further affects the passengers' expected waiting times. Again, it will influence the demand either through cancellation of the current requests or the conversion of future trip inquiries.
Because of its close ties to SD distributions, dynamic pricing is often jointly optimized with order matching or vehicle repositioning.  Within the (non-RL) operations research literature, dynamic pricing for ridesharing has already been studied and analyzed in conjunction with matching \citep{yan2020dynamic,ozkan2017dynamic} and from the spatiotemporal perspective \citep{ma2020spatio,bimpikis2019spatial,hu2021surge}, covering optimality and equilibrium analyses.

The complex interaction between pricing and the SD makes it hard to explicitly develop mathematical models that adapt well to dynamic and stochastic environments, and RL comes in as a promising direction to address these challenges by considering endogeneity and exogeneity as part of the environment dynamics.


Table \ref{tab:ref_pricing} summarizes the reviewed works on RL for dynamic pricing in ridesharing. 
As one of the early RL works, \cite{wu2016automated} consider a simplified ridesharing environment which captures only the two-sidedness of the market but not the spatiotemporal dimensions. The state of the MDP is the current price plus SD information. The action is to set a price, and the reward is the generated profit. A Q-learning agent is trained in a simple simulator, and empirical advantage in the total profit is demonstrated against other heuristic approaches.
More recent works leverage the spatiotemporal nature of the pricing actions and take into account the spatiotemporal long-term values in the pricing decisions.
\cite{chen2019inbede} integrate contextual bandits and the spatiotemporal value network developed in \citep{tang2019deep} for matching to jointly optimize pricing and matching decisions. In particular, the pricing actions are the discretized price percentage changes and are selected by a contextual bandits algorithm, where the long-term values learned by the value network are incorporated into the bandit rewards. In \citep{turan2020dynamic}, the RL agent determines both the price for each origin-destination (OD) pairs and the reposition/charging decisions for each electric vehicle in the fleet. The state contains global information such as the electricity price in each zone, the passenger queue length for OD pair, and the number of vehicles in each zone and their energy levels. The reward accounts for trip revenue, penalty for the queues, and operational cost for charging and repositioning. Due to the multi-dimensional continuous action space, PPO is used to train the agent in a simulator. \cite{song2020application} perform a case study of ridesharing in Seoul. They use a tabular Q-learning agent to determine spatiotemporal pricing, and extensive simulations are performed to analyze the impact of surge pricing on alleviating the problem of marginalized zones (areas where it is consistently hard to get a taxi) and on improving spatial equity.
\cite{chen2021spatial} adopt PPO to optimize the spatiotemporal pricing decisions for each hexagonal cell in terms of the per-km rate for the excess mileage beyond a base trip distance and the per-km rate for driver wage, for the objective of maximizing profits (revenue minus wage). The agent is modeled as a global decision-maker with state information of the numbers of open requests, vacant vehicles, occupied vehicles in each grid cell at time $t$ and historical demand at time $t-1$.
Unlike the works above that focus on the pricing decisions, \cite{mazumdar2017gradient} study from a different perspective of the pricing problem. The proposed risk-sensitive inverse RL method \citep{ng2000algorithms} recovers the policies of different types of passengers (risk-averse, risk-neutral, and risk-seeking) in view of surge pricing. The policy determines whether the passenger should wait or take the current ride.

As discussed in Section \ref{sec:scope}, under the setting where the driver pay is associated with the trip fare, the dynamic pricing policy also affects supply elasticity, i.e., drivers' decisions on participation in a given marketplace, working hours, and in some cases, the probability of accepting a given assignment, depending on the rules of the particular ridesharing platform \citep{chen2016dynamic,sun2019model,angrist2021uber}. Although not yet being widely considered in RL approaches to pricing, supply elasticity is an important piece of system state information that has significant implication to the sequence of pricing decisions. For the closely related topic of driver incentives design, \cite{shang2019environment,shang2021partially} adopts a learning-based approach to construct a generative model for driver behavior with respect to the incentives policy and subsequently trains an RL agent to optimize the incentives design for system-level metrics. Perhaps this example sheds some light on how RL is able to help improve pricing policies in view of supply-side effects.

\begin{table}
\hspace{-0.15\textwidth}\small
\begin{tabular}{||p{0.2\textwidth}|p{0.15\textwidth}|p{0.15\textwidth}|p{0.15\textwidth}|p{0.15\textwidth}|p{0.15\textwidth}|p{0.15\textwidth}||} 
\hline
Paper & Agent & State & Action & Reward & Algorithm & Environment \\
\hline\hline
\cite{wu2016automated} & global decision-maker & current trip price (same for all trips), SD info & price & profit & Q-learning & no spatiotemporal dimensions \\
\hline
\cite{chen2019inbede} & global decision-maker & features of the trip request & discretized price change percentage & profit & contextual bandits with action values partly computed by CVNet & ride-hailing simulator with pricing module and passenger elasticity model \\
\hline
\cite{turan2020dynamic} & global decision-maker for pricing and EV charging & electricity price in each zone, passenger queue length for each OD pair, number of vehicles in each zone and their energy levels & price for each OD pair, reposition/ charging for each vehicle & trip revenue - penalty for queues - operational cost for charging and reposition & PPO & simulator \\
\hline
\cite{song2020application} & global decision-maker & location, time & price for spatialtemporal grid cells & trip price minus penalty for driver waiting & Q-learning & case study: ride-hailing simulation of Seoul \\
\hline
\cite{mazumdar2017gradient} & passenger & price multiplier, time, if a ride has completed & wait, take current ride & trip price to pay & risk-sensitive inverse RL & historical data \\
\hline
\cite{chen2021spatial} & global decision-maker & number of open requests, vacant vehicles, and occupied vehicles in each grid cell at time $t$, and demand in time $t-1$ & joint actions of price (per-km for excess mileage) and wage (per-km rate) for each grid cell & profit: revenue minus wage & PPO & simulation based on Hangzhou data from DiDi; modeling on both supply and demand elasticity \\
\hline\hline
\end{tabular}
\caption{Summary of literature for Pricing.}
\label{tab:ref_pricing}
\end{table}

\subsection{Online Matching}\label{sec:matching}
\begin{figure}
\begin{center}
        \includegraphics[width=0.9\linewidth]{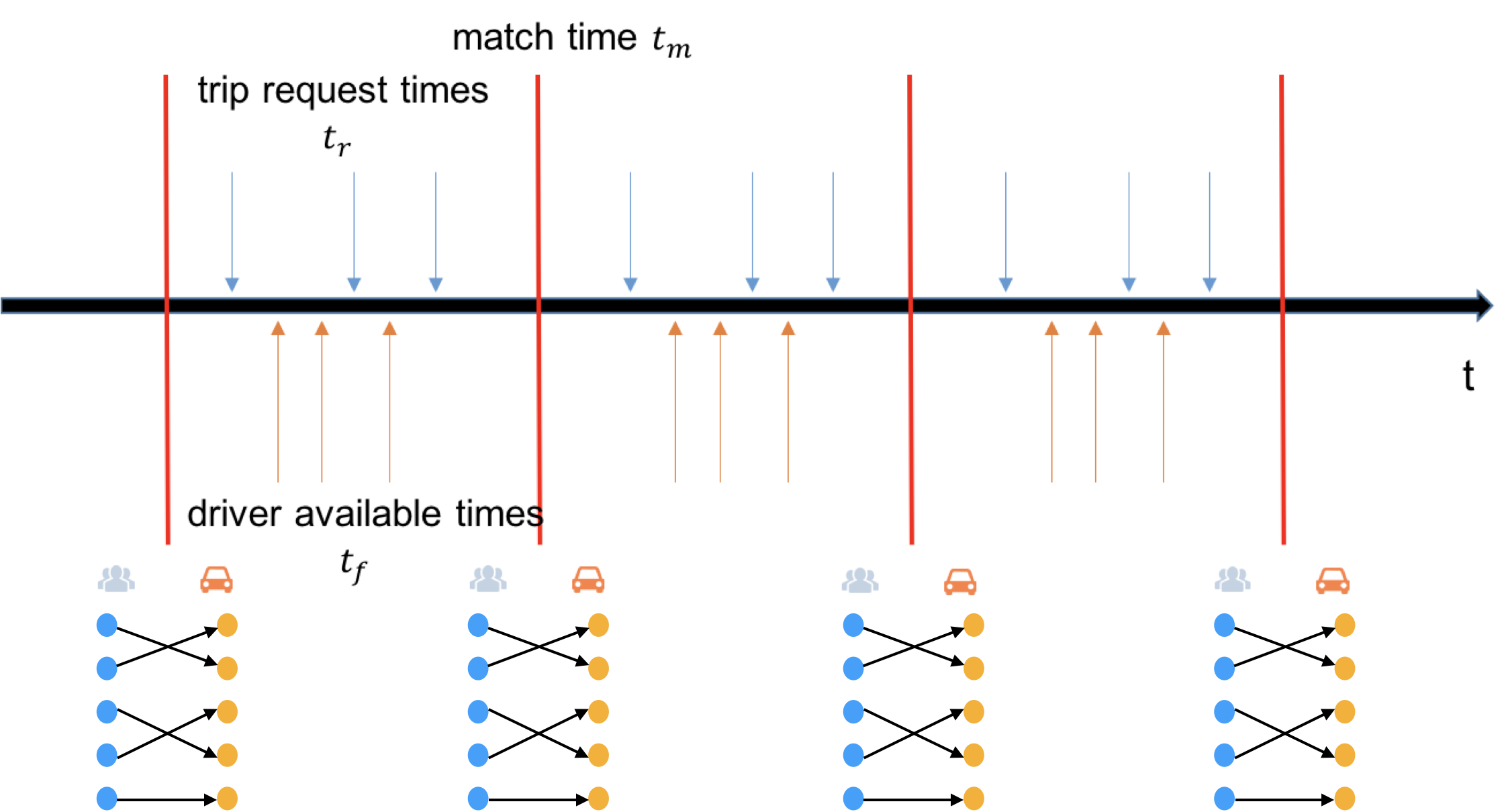}
	\caption{The order matching process with batching from the system perspective \citep{qin2020ride}. The assignments are for illustration only.}
	\label{fig:batch_match}
\end{center}
\end{figure}

The rideshare matching problem and its generalized forms have been investigated extensively in the field of operations research (see e.g., \citep{ozkan2017dynamic,hu2022dynamic,lowalekar2018online} and the references therein). Typically, both the open trip requests and available drivers are batched within time windows of fixed length as they arrive at the system, and they are matched at predefined discrete review times. See Figure \ref{fig:batch_match} for an illustration. Hence, ridesharing matching is an online stochastic problem \citep{qin2020ride}.
Outside the RL literature, \cite{lowalekar2018online} approach the problem through stochastic optimization and use Bender's decomposition to solve it efficiently. To account for the temporal dependency of the decisions, \cite{hu2022dynamic} formulate the problem as a stochastic DP and propose heuristic policies to compute the optimal matching decisions. 
For a related problem, the truckload carriers assignment problem, \cite{simao2009approximate} also formulate a dynamic DP but with post-decision states so that they are able to solve the problem using ADP. In each iteration, a demand path is sampled, and the value function is approximated in a linear form and updated using the dual variables from the LP solution to the resulting optimization problem.  

\begin{figure}
\begin{center}
        \includegraphics[width=0.9\linewidth]{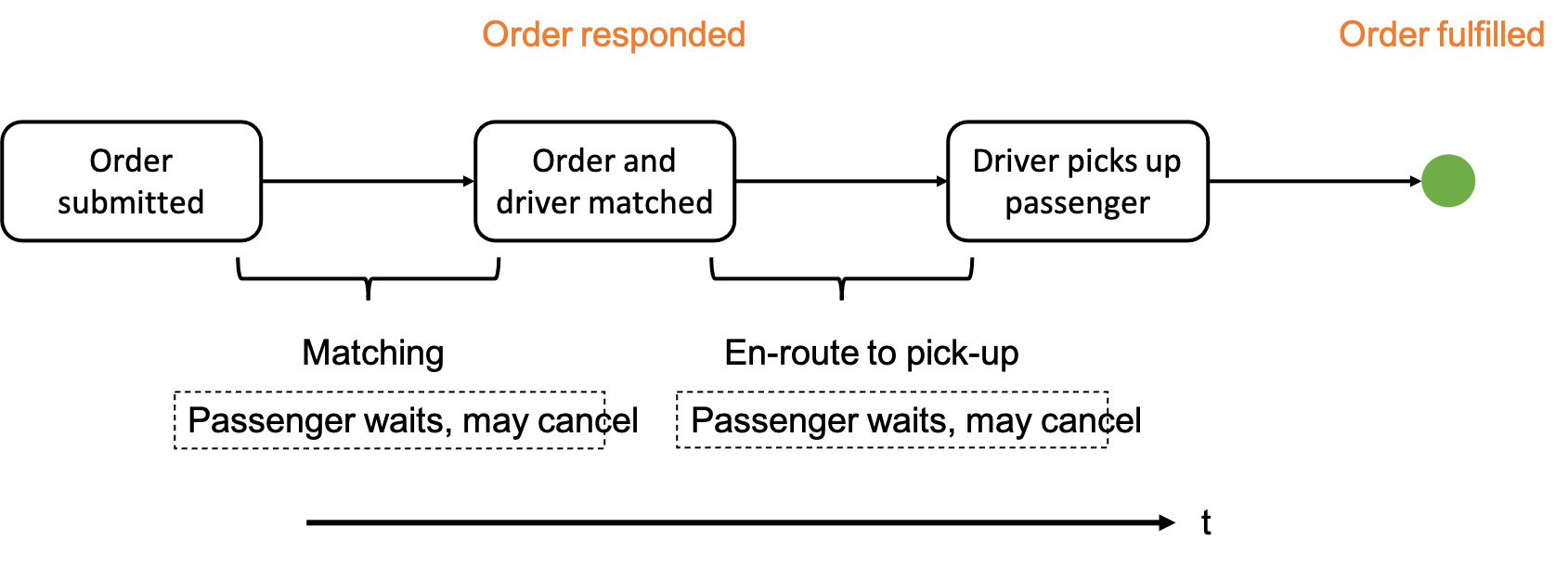}
	\caption{The order matching process from a single request's perspective.}
	\label{fig:matching_process}
\end{center}
\end{figure}

The RL literature for rideshare matching (see Table \ref{tab:ref_matching}) typically aims to optimize the total driver income and the service quality over an extended period of time. Service quality can be quantified by \emph{response rate} and \emph{fulfillment rate}. Response rate is the ratio of the matched requests to all trip requests. Since the probability of pre-match cancellation is primarily a function of response time (pre-match waiting time), the total response time is an alternative metric to response rate. Fulfillment rate is the ratio of completed requests to all requests and is no higher than the response rate. The gap is due to post-match cancellation, usually because of the waiting for pick-up. 
Hence, the average pick-up distance is also a relevant quantity to observe. Figure \ref{fig:matching_process} shows the detailed flow of matching a single trip request together with the quantities discussed above.

In terms of the MDP formulation, driver agent is a convenient modeling choice for its straightforward definition of state, action, and reward, in contrast to system-level modeling where the action space is exponential. In this case, the rideshare platform is naturally a multi-agent system with a global objective. A common approach is to crowdsource all drivers' experience trajectories to train a single agent and apply it to all the drivers to generate their matching policies \citep{xu2018large,qin2018dispatching,tang2019deep}. Since the system reward is the sum of the drivers' rewards, the system value function does decompose into the individual drivers' value functions computed by each driver's own trajectories. The approximation here is using a single value function learned from all drivers' data. See \citep{qin2020ride} for detailed discussions. Specifically, \cite{xu2018large} learn a tabular driver value function using TD(0), and \cite{qin2018dispatching,tang2019deep,holler2019deep} apply DQN-type of training to learn a value network. In particular, \cite{tang2019deep} design a spatiotemporal state-value network using hierarchical coarse coding and cerebellar embedding memories for better state representation and training stability. \cite{holler2019deep} develop an action-value network that leverages global SD information, which is embedded into a global context by attention.

This type of single-agent approach avoids dealing explicitly with the multi-agent aspect of the problem and the interaction among the agents during training. Besides simplicity, this strategy has the additional advantage of being able to easily handle a dynamic set of agents (and hence, a changing action space) \citep{ke2020learning}. On the other hand, order matching requires strong system-level coordination in that a feasible solution has to satisfy the one-to-one constraints. To address this issue, \cite{xu2018large,tang2019deep} use the learned state values to populate the edge weights of a bipartite assignment problem to generate a collective-greedy policy \citep{qin2020ride} with respect to the state values.  \cite{holler2019deep} assume a setting where drivers are matched or repositioned sequentially so that the policy output always satisfies the matching constraints.

Leveraging MARL, \cite{li2019efficient,jin2019coride,zhou2019multi} directly optimize the multi-agent system. One significant challenge is scalability since any realistic ridesharing setting can easily involve thousands of agents, precluding the possibility of dealing with an exact joint action space. \cite{li2019efficient} apply mean-field MARL to make the interaction among agents tractable, by taking the `average' action of the neighboring agents to approximate the joint actions. \cite{zhou2019multi} argue that no explicit communication among agents is required for order matching due to the asynchronous nature of the transitions and propose independent Q-learning with centralized KL divergence (of the supply and demand distributions) regularization. Both \cite{li2019efficient,zhou2019multi} follow the centralized training decentralized execution paradigm. \cite{jin2019coride} take a different approach treating each spatial grid cell as a worker agent and a region of a set of grid cells as a manager agent, and they adopt hierarchical RL to jointly optimize order matching and vehicle repositioning.

In practical settings, the online matching policy often has to balance among multiple objectives \citep{lyu2019multi}, e.g., financial metrics and customer experience metrics. The rationale is that persistent negative customer experience will eventually impact long-term financial metrics as users churn the service or switch to competitors. There are two potential ways that one can leverage RL to approach this problem. The explicit approach is to directly learn a policy that dynamically adjusts the weights to combine the multiple objectives into a single reward function. With the abundance of historical experience data, inverse RL can be used to learn the relative importance of multiple objectives under a given unknown policy \citep{zhou2021multi}. The implicit approach is to capture the necessary state signals that characterize the impact of the metrics not explicitly in the reward function, so that the learned value function correctly reflect the long-term effect of the multiple metrics. As discussed in Section \ref{sec:behavior}, the long feedback loop is a potential challenge here.

Besides the driver-passenger pairing decisions, there are other important levers that can be optimized within the matching module, namely the matching window and the matching radius \citep{yang2020optimizing}. The matching window determines when to match a request (or a batch of requests). A larger window increases pre-match waiting time but may decrease pick-up time for matched requests because of more available drivers. There have been several RL works on the matching window optimization, which can be done from the perspective of a request itself \citep{ke2020learning} or the system \citep{wang2019adaptive,qin2021optimizing}. In \citep{ke2020learning}, each trip request is an agent. An agent network is trained centrally using pooled experience from all agents to decide whether or not to delay the matching of a request to the next review window, and all the agents share the same policy. To encourage cooperation among the agents, a specially shaped reward function is used to account for both local and global reward feedback. They also modify the RL training framework to address the delayed reward issue by sampling complete trajectories at the end of training epochs to update the network parameters. \cite{wang2019adaptive} take a system's view and 
propose a Restricted Q-learning algorithm to determine the length of the current review window (or batch size). They show theoretical analysis results on the performance guarantee in terms of competitive ratio for dynamic bipartite graph matching with adaptive windows. \cite{qin2021optimizing} take a similar modeling perspective but use the AC method with experience replay (ACER) \citep{wang2016sample} that combines on-policy updates (through a queuing-based simulator) with off-policy updates. The matching radius defines how far an idle driver can be from the origin of a given request to be considered in the matching. It can be defined in travel distance or time. A larger matching radius may increase the average pick-up distance but requests are more likely to be matched within a batch window, whereas a smaller radius renders less effective driver availability but it may decrease the average pick-up distance. Both the matching window and radius are trade-offs between pre-match and post-match waiting times (and hence, cancellation).
So far, few effort through RL has been devoted to matching radius optimization. The joint optimization of the matching window and radius is certainly another interesting line of research.


Because of its generalizability, matching for ridesharing is closed related to a number of online matching problems in other domains, the RL methods to which are also relevant and can inform the research in rideshare matching. Some examples are training a truck agent using DQN with pooled experience to dispatch trucks for mining tasks \citep{zhang2020dynamic}, learning a decentralized value function using PPO with a shaped reward function for cooperation (in similar spirit as \citep{ke2020learning}) to dispatch couriers for pick-up services \citep{chen2019can}, and designing a self-attention, pointer network-based policy network for a system agent to assign participants to tasks in mobile crowdsourcing \citep{shen2020auxiliary}.

\begin{sidewaystable}
\scriptsize
\begin{tabular}{||p{0.125\textwidth}|p{0.05\textwidth}|p{0.1\textwidth}|p{0.125\textwidth}|p{0.175\textwidth}|p{0.125\textwidth}|p{0.125\textwidth}|p{0.1\textwidth}|p{0.125\textwidth}||} 
\hline
Paper & Agent & State & Action & Reward & Algorithm & Environment & Notes \\
\hline\hline
\cite{xu2018large} & driver & assignment to a specific order or idle & location, time & trip price & tabular TD(0) for learning state values offline + Hungarian method for generating the assignment online & deployed in production;
multi-agent, agent-level simulation & \\
\hline
\cite{qin2018dispatching} & driver & assignment to a specific order or idle & location, time, SD features & trip price & offline DQN for matching,
CFPT network for transfer learning & single-agent simulation & single-vehicle problem \\
\hline
\cite{tang2019deep}, \cite{qin2020ride} & driver & assignment to a specific order or idle & location, time, SD features & trip price & CVNet (deep TD-like algorithm) for learning state values offline + Hungarian method for generating the assignment online & deployed in production;
multi-agent, agent-level simulation & hierarchical sparse coarse coding, cerebellar embedding of spatial info, Lipschitz regularization on network \\
\hline
\cite{holler2019deep} & driver, system & matching a driver to an order, reposition a driver;
matching and repositioning done sequentially & global info of all drivers and open orders & trip price, reposition cost & DQN, PPO & multi-agent, agent-level simulation & attention mechanism to extract global state info into a context vector \\
\hline
\cite{li2019efficient} & driver & assignment to a specific order or idle & location, time, is\_available & trip price & mean-field MARL, AC method & homogeneous vehicles within the same grid cell & The mean action is represented by the peers' destination distribution. \\
\hline
\cite{jin2019coride} & worker: hex cell
manager: group of hex cells (one layer) & worker: ranking for match and reposition

manager: abstract goal for workers & number of vehicles, orders, entropy, reposition-guided vehicles, distributions of trip prices and durations in the given hex cell & manager: total driver income + specifically designed quantity to promote high order response rate

worker: intrinsic reward for following the goal generated by manager & hierarchical MARL & homogeneous vehicles within the same grid cell & multiple managers, each manager communicates with multiple workers

multi-head attention mechanism for coordination \\
\hline
\cite{zhou2019multi} & driver & a trip tuple: (origin cell, destination cell, trip duration, price) & cell index, number of idle vehicles, orders, distribution of trip destinations in the given hex cell & trip price & independent learning with KL divergence regularization & homogeneous vehicles within the same grid cell & \\
\hline
\cite{ke2020learning} & trip request & match or delay & global: number of idle vehicles, open requests, expected arrival rates of requests and drivers in each cell

local: location, cumulative waiting time, expected pick-up distance & local reward based on the ultimate outcome of matching (whether or not matched or cancelled): trip value, pick-up distance, and match window time

Global reward is based on average local reward.

Final reward is convex combination of local and global rewards.

The rewards have to be updated at the end of the epoch. & DQN, PPO, A2C, ACER with delayed reward. Whole episode trajectories are sampled from replay buffer. & agent-based simulation with delayed matching feature & \\
\hline
\cite{wang2019adaptive} & system & expected length of current batch & current nodes in the bipartite graph, current batch size & the sum of the edge weights in the batch & restricted Q-learning & DiDi GAIA data set & The adaptive batch-based matching has a guarantee on competitive ratio \\
\hline
\cite{qin2021optimizing} & system & match the current batch or continue to batch (decision made at every time interval) & number of batched requests and vehicles and estimated arrival rates of demand and supply in each cell & for each time interval, the negative of total matching wait time for all batched requests and the total pick-up wait time saved (by delaying the current batch) & ACER that combines on-policy updates (through a queuing-based simulator) with off-policy updates & Shanghai taxi data & \\
\hline
\cite{shi2019operating} & vehicle & remaining battery level when available, next available time and location, global time & matching, EV (charging) & trip price - pick-up cost - charging cost & similar to CVNet \citep{tang2019deep} & synthetic data & assumes decomposibility of system value into vehicle values \\
\hline
\cite{kullman2021dynamic} & system & global time, new request info, state of each vehicle & joint matching, reposition, and charging (EV) for each vehicle & revenue - travel cost & DQN + attention mechanism over vehicles embeddings (similar to \citep{holler2019deep}) & NYC taxi data with taxi zones & Decision epoch is either a new request arrives or a vehicle becomes idle. \\
\hline
\cite{al2020approximate} & system & supply-demand counts in spatiotemporal discretized space & joint matching and reposition, EV (charging) & revenue - charging expense & ADP with value function approx. on post-decision states.
Value function approx. by hierarchical aggregation. & simulation with New Jersey ride-hailing trip data \\
\hline
\hline
\end{tabular}
\caption{Summary of literature for Online Matching.}
\label{tab:ref_matching}
\end{sidewaystable}

\subsection{Vehicle Repositioning}\label{sec:reposition}
Vehicle repositioning from a single-driver perspective (i.e., taxi routing) has a relatively long history of research
since taxi service has been in existence long before the emergence of rideshare platforms. Likewise, research on RL-based approaches for this problem (see Table \ref{tab:ref_reposition_taxi}) also appeared earlier than that on system-level vehicle repositioning. 

\begin{figure}
\begin{center}
    \includegraphics[width=0.5\linewidth]{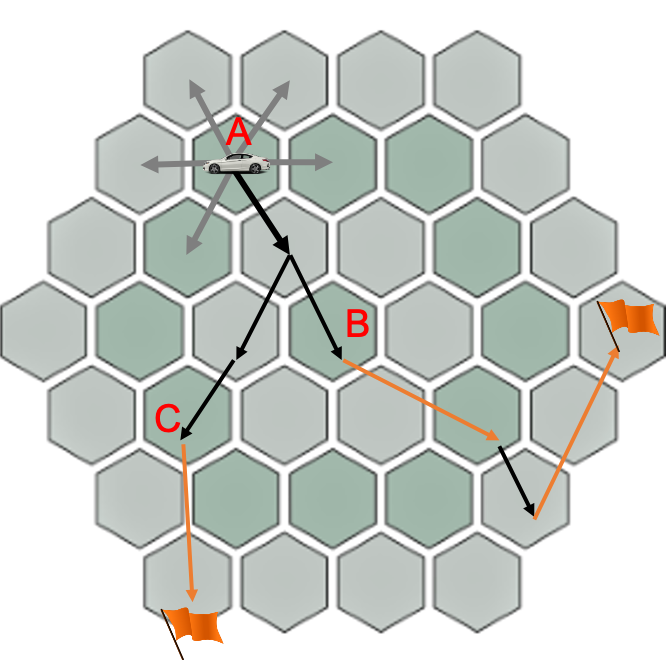}
	\caption{Illustration of the (single-agent) taxi routing problem on a hexagon grid system. The vehicle at its origin position A has the option to reposition to one of the neighboring and current cells. The black arrows represent reposition (idle cruising), and in the two scenarios, the vehicle is matched to a trip request at B and C respectively. The orange arrows represent trip moves, and the orange flags are where the episodes terminate (for long-term horizons).}
	\label{fig:reposition_single}
\end{center}
\end{figure}

For the taxi routing problem, each driver is naturally an agent, and the objective thus focuses on optimizing individual reward. Common reward definitions include trip fare \citep{rong2016rich}, net profit (income - operational cost) \citep{verma2017augmenting}, idle cruising distance \citep{garg2018route}, and ratio of trip mileage to idle cruising mileage \citep{gao2018optimize}.
Earlier works \cite{han2016routing,wen2017rebalancing,verma2017augmenting,garg2018route} optimize the objective within a horizon up to the next successful match (i.e., A-to-B and A-to-C in Figure \ref{fig:reposition_single}), but it is now more common to consider a long-term horizon, where an episode usually consists of a trajectory over one day \citep{lin2018efficient,shou2020optimal,jtq2021repos}. We illustrate these concepts in Figure \ref{fig:reposition_single}.

The type of actions of an agent depends on the physical abstraction adopted. A simpler and more common way of representing the spatial world is a grid system, square or hexagonal\footnote{The hexagonal grid system is the industry standard.} \citep{han2016routing,wen2017rebalancing,verma2017augmenting,gao2018optimize,lin2018efficient,rong2016rich,jtq2021repos,shou2020optimal}. In this setting, the action space is the set of neighboring cells (often including the current cell). \cite{shou2020reward} explain the justification for this configuration. Determination of the specific destination point is left to a separate process, e.g., pick-up points service \citep{jtq2021repos}. The more realistic abstraction is a road network, in which the nodes can be intersections or road segments \citep{garg2018route,yu2019markov,zhou2018optimizing,schmollsemi}. The action space is the adjacent nodes or edges of the current node. This approach supports a turn-by-turn guiding policy but requires more map information at run time. 

Most of the papers adopt a tabular value function, so the state is necessarily low-dimensional, including spatiotemporal information and sometimes additional categorical statuses. \cite{shou2020optimal} have a boolean in the state to indicate if the driver is assigned to consecutive requests since its setting allows a driver to be matched before completing a trip request. \cite{rong2016rich,zhou2018optimizing} have the direction from which the driver arrives at the current location. 
For deep RL-based approaches \citep{wen2017rebalancing,jtq2021repos}, richer contextual information, such as SD distributions in the neighborhood, can go into the state.

The learning algorithms are fairly diverse but are all value-based. By estimating the various parameters (e.g., matching probability, passenger destination probability) to compute the transition probabilities, \cite{rong2016rich,yu2019markov,shou2020optimal,zhou2018optimizing} adopt a model-based approach and use value iterations to solve the MDP. 
\cite{shou2020optimal} further use inverse RL to learn the unit-distance operational cost.
Model-free methods are also common, e.g., Monte Carlo learning \citep{verma2017augmenting}, Q-learning \citep{han2016routing,gao2018optimize}, and DQN \citep{wen2017rebalancing}. \citep{jtq2021repos} is a hybrid approach in that it performs an action tree search at the online planning stage using estimated matching probabilities and a separately learned state value network. \citep{garg2018route} is in a similar spirit by augmenting the multi-arm bandits with Monte Carlo tree search.

\begin{figure}
\begin{center}
    \includegraphics[width=0.7\linewidth]{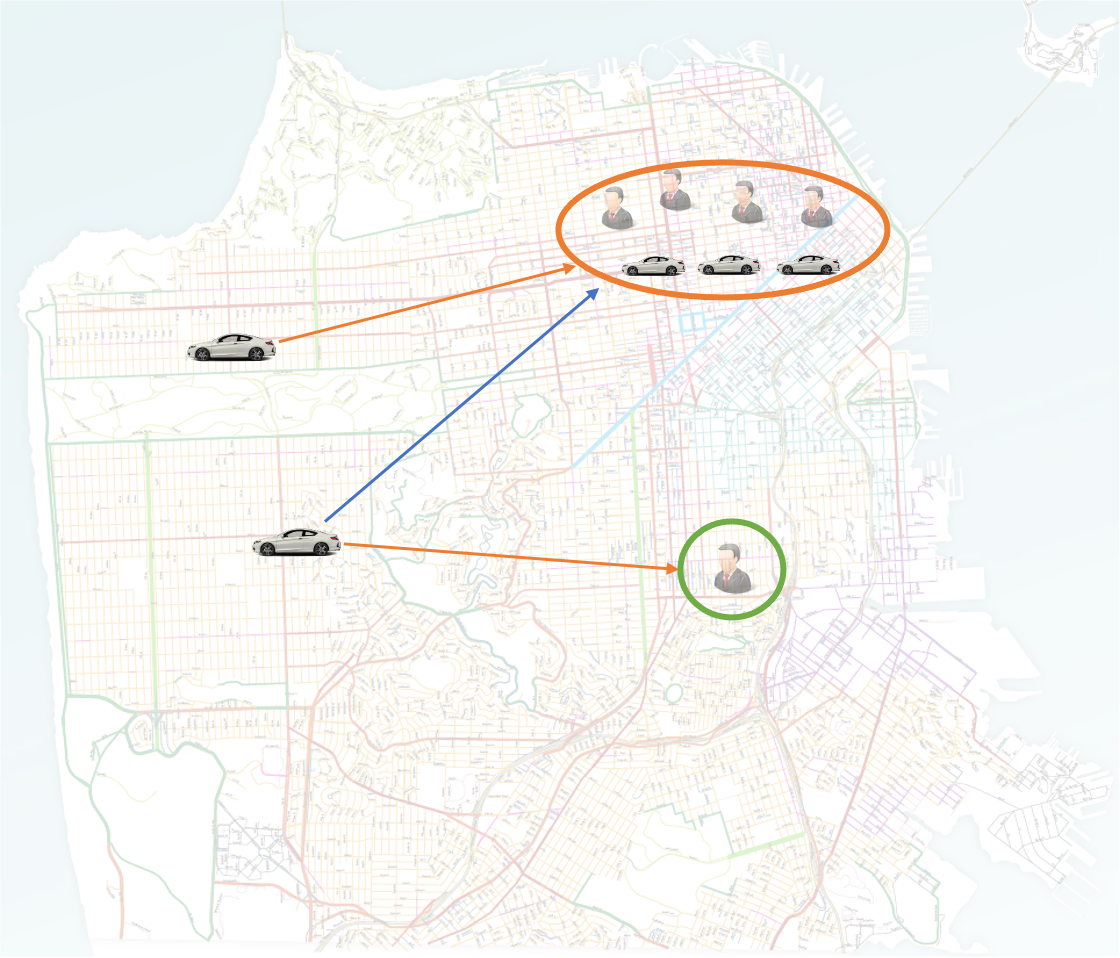}
	\caption{Illustration of system-level vehicle repositioning. The requests in the orange-circled and green-circled areas appear in the future w.r.t. the time of repositioning. The empty vehicles are existing ones in the orange-circled area. The orange and blue arrows represent potential reposition moves.}
	\label{fig:reposition_system}
\end{center}
\end{figure}

The problem formulation most relevant to the ridesharing service provider is system-level vehicle repositioning. Similar to order matching, the ridesharing platform reviews the vehicles' states at fixed time intervals which are significantly longer than those for order matching. See Figure \ref{fig:batch_match} for illustration. Idle vehicles that meet certain conditions, e.g., being idle for sufficiently long time and not in the process of an existing reposition task, are sent reposition recommendations, which specify the desired destinations and the associated time windows. 
The motivation here is to explicitly modify the current distribution of the available vehicles so that collectively they are better positioned to fulfill more requests more efficiently in the future. Figure \ref{fig:reposition_system} explains the idea with a concrete example. If the vehicles reposition independently (following the orange arrows), they both move to the orange-circled area and there will be a surplus of supply while the demand in the green-circled area will not be served. In contrast, if the vehicles coordinate and the one in the south repositions by the blue arrow, both vehicles will be matched, and all the requests are served.

The agent can be either the platform or a vehicle, latter of which calls for a MARL approach. All the works in this formulation have global SD information (each vehicle and request's status or SD distributions) in the state of the MDP, and a vehicle agent will additionally have its spatiotemporal status in the state. The rewards are mostly the same as in the taxi routing case, except that \cite{mao2020dispatch} consider the monetized passenger waiting time. The actions are all based on grid or taxi zone systems.

The system-agent RL formulation has only been studied very recently, in view of the intractability of the joint action space of all the vehicles (see Table \ref{tab:ref_reposition_sys}). To tackle this challenge of scalability, \cite{feng2020scalable} decompose the system action into a sequence of atomic actions corresponding to passenger-vehicle matches and vehicle repositions. The MDP encloses a `sequential decision process' in which all feasible atomic actions are executed to represent one system action, and the MDP advances its state upon complete of the system action. They develop a PPO algorithm for the augmented MDP to determine the sequence of the atomic actions. The system policy in \citep{mao2020dispatch} produces a reposition plan that specifies the number of vehicles to relocate from zone $i$ to $j$ so that the action space is independent from the number of agents (at the expense of additional work at execution). The agent network, trained by a batch AC method, outputs a value for each OD pair, which after normalization gives the percentage of vehicles from each zone to a feasible destination. 

The vehicle-agent approaches have to address the coordination issue among the agents. 
\cite{lin2018efficient} develop contextual DQN and AC methods, in which coordination is achieved by masking the action space based on the state context and splitting the reward accrued in a grid cell among the multiple agents within the same cell. 
\cite{oda2018movi} treat the global state in grid as image input and develop an independent DQN method. They argue that independent learning, equipped with global state information, works quite well compared to an MPC-based approach.
The zone structure in \citep{liu2020context} is constructed by clustering a road-connectivity graph. A single vehicle agent is trained with contextual deep RL and generates sequential actions for the vehicles.
\cite{zhang2020dynamic} also train a single DQN agent for all agents, but with global KL distance between the SD distributions similar to \citep{zhou2019multi}. The DQN agent is put in tandem with QRewriter, another agent with a Q-table value function that converts the output of DQN to an improved action.
\cite{shou2020reward} approach the MARL problem with bilevel optimization: The bottom level is a mean-field AC method \citep{li2019efficient} with the reward function coming from a platform reward design mechanism, which is tuned by the top level Bayesian optimization.
Agent coordination is done by a central module in \citep{chaudhari2020learn}, where a vehicle agent executes a mix of independent and coordinated actions. The central module determines the need for coordination based on SD gaps, and explicit coordination is achieved by solving an assignment problem to move vehicles from excess zones to deficit zones.

For joint matching and repositioning optimization, one major challenge is the heterogeneous review cadence. Matching and reposition decisions are typically made asynchronously in practice. To address this issue, \cite{tang2021value} allow the two modules to operate independently but share the same spatiotemporal state value function which is updated online. If the two decisions are formulated into the same problem, the action space can be masked depending on the state \citep{holler2019deep}. 

Existing RL literature on repositioning often assumes the drivers' full compliance to reposition, i.e., the autonomous vehicle setting. How non-compliance affects the overall performance of a reposition algorithm is a natural question to ask when considering a real-world ridesharing system, in which we expect to see a combination of drivers' independent cruising strategies \citep{urata2021learning,wong2014cell} and system-guided idle cruising behavior. It is also interesting and practically necessary to investigate incentives design and strategies that facilitate the repositioning process. In \citep{zhu2021mean}, for example, a mean-field MDP is developed for modeling drivers' strategies, and empirical investigations are performed on how spatiotemporal driver incentives affect driver behavior and the system performance.

\begin{sidewaystable}
\scriptsize
\begin{tabular}{||p{0.1\textwidth}|p{0.025\textwidth}|p{0.025\textwidth}|p{0.1\textwidth}|p{0.125\textwidth}|p{0.125\textwidth}|p{0.075\textwidth}|p{0.125        \textwidth}|p{0.1\textwidth}|p{0.075\textwidth}||} 
\hline
Paper & Type & Agent & State (in addition to ST info) & Action & Reward & Episode & Algorithm & Coordination & Data \\
\hline\hline
\cite{rong2016rich} & taxi & driver & direction from which driver arrives at the current location & neighboring cells in a grid system & trip fare & long-term & model-based, VI to solve MDP & - & \\
\hline
\cite{han2016routing} & taxi & driver & - & neighboring cells in a grid system & trip fare - reposition cost & from idle to completion of next trip(s) and being idle again & Q-learning & - & \\
\hline
\cite{verma2017augmenting} & taxi & driver & - & neighboring cells in a grid system & trip fare - reposition cost & up to the next match & MC learning & - & taxi log data from Singapore \\
\hline
\cite{wen2017rebalancing} & taxi & driver & SD contextual info & neighboring cells in a grid system & saved idle time compared to a counterfactual simulation without reposition. Penalized if no match after reposition. & up to the next match & DQN with greedy action

Compared with MINLP and SAR (simple anticipatory rebalancing) on avg passenger wait time. & - & \\
\hline
\cite{garg2018route} & taxi & driver & current node on road network & adjacent node or edge & idle cruising distance till the next passenger & up to the next match & MAB + MCTS & - & \\
\hline
\cite{gao2018optimize} & taxi & driver & location, {occupied, parking, vacant} & idle cruise to a neighboring cell in the grid system, carrying passenger to destination, waiting & trip mileage/idle cruising mileage

Problem objective: 
effective driving ratio = total trip mileage / total idle cruising mileage & long-term (day) & Q-learning & - & Beijing taxi data 2013 \\
\hline
\cite{zhou2018optimizing} & taxi & driver & current road segment, time, previous road segment & adjacent node or edge & trip fare
 & long-term & model-based, VI to solve MDP & - & a year of taxi data from a major city in China \\
 \hline
 \cite{yu2019integrated} & taxi & driver & current node on road network & adjacent node or edge & trip fare - operational cost & long-term (day) & model-based, VI to solve MDP;
use parallel matrix operations to accelerate computation & - & Shanghai taxi trajectory data in the morning of a weekday \\
\hline
\cite{shou2020optimal} & taxi, system & driver & boolean: whether or not assigned to consecutive requests & neighboring cells in a grid system & trip fare - operational cost

operational cost per unit distance learned through IRL & long-term (day) & model-based, VI to solve MDP;
inverse RL to learn unit distance op cost & Sequentially make decision for each driver. Adjust the matching prob of each driver's MDP, and solve again. & Beijing ride-hailing trajectory data from DiDi over three weekdays \\
\hline
\cite{jtq2021repos} & taxi & vehicle & SD contextual info in the current cell & neighboring cells in a grid system & trip fare - reposition cost & long-term & offline CVNet + decision-time action search & - & DiDi ride-hailing data \\
\hline
\hline
\end{tabular}
\caption{Summary of literature for Vehicle Repositioning (taxi routing).}
\label{tab:ref_reposition_taxi}
\end{sidewaystable}

\begin{sidewaystable}
\scriptsize
\begin{tabular}{||p{0.1\textwidth}|p{0.025\textwidth}|p{0.025\textwidth}|p{0.1\textwidth}|p{0.125\textwidth}|p{0.125\textwidth}|p{0.075\textwidth}|p{0.125        \textwidth}|p{0.1\textwidth}|p{0.075\textwidth}||} 
\hline
Paper & Type & Agent & State (in addition to ST info) & Action & Reward & Episode & Algorithm & Coordination & Data \\
\hline
\hline
\cite{lin2018efficient} & system & driver & global SD contextual features in all cells & neighboring cells in a grid system & trip fare, shared when multiple agents are in the same grid cell & long-term & contextual DQN, AC & contextual state features, action space pruned by context & 4 weeks of DiDi data in Chengdu, China \\
\hline
\cite{oda2018movi} & system & driver & global SD state discretized into cells, treated as an image & reachable cells in a grid system within the reposition cycle & weighted number of pick-ups - reposition time & long-term & independent DQN & - & NYC taxi data \\
\hline
\cite{shou2020reward} & system & vehicle & - & neighboring cells in a grid system & trip fare & long-term & bilevel optimization: top Bayesian optimization to update reward param, bottom mean-field MARL (AC) & mean-field MARL & NYC taxi data \\
\hline
\cite{jtq2021repos} & system & vehicle & SD contextual info in the current and neighboring cells & neighboring cells in a grid system & trip fare - reposition cost & long-term & deep SARSA & stochastic policy through softmax of action values, SD regularization to action values & DiDi ride-hailing data \\
\hline
\cite{mao2020dispatch} & system & system & SD info for each zone & reposition plan: number of repositioned vehicles for each OD pair in a zone map & monetized passenger waiting time & long-term & batch AC  & central decision-making & NYC taxi data \\
\hline
\cite{feng2020scalable} & system & system & status of every vehicle and request & atomic action: driver-passenger match or driver-destination match (reposition);
system action: sequence of atomic actions & trip fare - operational cost & long-term (day) & PPO applied to MDP with Sequential Decision Process embedded: global actions decomposed into sequential atomic ones & central decision-making & DiDi data: 5 regions, 1000 cars, 360 minutes \\
\hline
\cite{liu2020context} & system & vehicle & discrete zone structure constructed by clustering a road connectivity graph & neighboring zones & trip fare & long-term & contextual DQN with shared value function & vehicle actions generated sequentially & real-world taxi data \\
\hline
\cite{zhang2020dynamic} & system & vehicle & global SD distributions & neighboring cells in a grid system & global KL distance between SD distributions & long-term & DQN + Q-table in tandem with shared value function & global state features & DiDi data \\
\hline
\cite{chaudhari2020learn} & system & vehicle & cell in an ST table & neighboring cells in a grid system, wait & trip fare - operational cost & long-term & SARSA-like policy evaluation & solve an assignment problem of surplus and deficits in terms of SD gap & NYC taxi data \\
\hline
\hline
\end{tabular}
\caption{Summary of literature for Vehicle Repositioning (system reposition).}
\label{tab:ref_reposition_sys}
\end{sidewaystable}

\subsection{Route Guidance (Navigation)}
Routing in this paper refers to low-level navigation decisions on a road network, typically with output of matching and repositioning algorithms as input. The road network, combined with traffic conditions on the links (exhibited as link costs), forms the traffic network which is a non-stationary stochastic network \citep{mao2018reinforcement}. It is known that standard static shortest-path algorithms do not find the path with minimum expected cost in this case, and the optimal route is not a simple route but a policy \citep{hall1986fastest,kim2005optimal}. There are two types of set-up for the routing problem, depending on the decision review time. In the first type of set-up, each vehicle on the road network selects a route for a given OD pair from a set of feasible routes. The decision is only reviewed and revised after a trip is completed. Hence, it is called route planning or route choice. When the routes for all the vehicles are planned together, it is equivalent to assigning the vehicles to each link in the network, and hence, the problem is called traffic assignment problem (TAP), which is typically for transportation planning purposes. In the second type of set-up, the routing decision is made at each intersection to select the next outbound road (link) to enter. These are real-time adaptive navigation decisions for vehicles to react to the changing traffic state of the road network. The problem corresponding to this set-up is called dynamic routing, dynamic route choice, or route guidance.

Routing on a road network is a typical multi-agent problem, where the decisions made by one agent has influence on the other agents' performance, simply in that the congestion level of a link depends directly on the number of vehicles passing through that link at a given time and has direct impact on the travel time for all the vehicles on that link within the same time interval.  
The literature for route planning and TAP often consider the equilibrium property of the algorithms when a population of vehicles adopt them. TAP is typically from a traffic manager's (i.e., system's) perspective. Its goal is to reach system equilibrium (SE, or also often referred to as the system optimum). Some works focus on route planning or TAP from an individual driver's perspective and maximize individual reward. These algorithms try to reach user equilibrium (UE) or Nash equilibrium, under which no agent has the incentive to change its policy because doing so will not achieve higher utility. This is the best that selfish agents can achieve but may not be optimal for the system. 

Value-based RL is by far the most common approach for route planning and TAP aiming to reach UE (see Table \ref{tab:ref_routing}). In the MDP formulation, the agent is a vehicle (or equivalently, a task) with a given OD pair. The objective is to minimize the total travel time for an individual vehicle (task) \citep{mainali2008optimal,ramos2018analysing,zhou2020reinforcement}, i.e., the agent is selfish. The immediate reward is the total travel time of a trip for an individual and a particular run. This MDP is stateless, so strictly speaking, it is a multi-arm bandits or contextual bandits problem \citep{li2010contextual} if considering time as a contextual feature.   The action to take at decision time is to select a route from the set of feasible routes for the associated OD pair \citep{ramos2018analysing,zhou2020reinforcement,bazzan2015hybrid}. The value function that governs the route choice decisions represents the long-term expected travel time for the trip identified by the given OD pair. Due to the multi-agent nature, the environment w.r.t. each agent is non-stationary in that the reward function is changing with the policy updates from the other agents. Empirical convergence to UE is demonstrated by \cite{ramos2018analysing}. \cite{zhou2020reinforcement} further develop a Bush-Mosteller RL scheme for MARL and formally establishes its UE convergence property. 
We also highlight some unique features of the papers. \cite{ramos2018analysing} consider a different objective from the common and minimizes the driver's regret. To do that, the Q-learning updates are modified using the estimated action regret, which can be computed by local observations and global travel time information communicated by an app. \cite{bazzan2015hybrid} propose a hybrid method, with Q-learning for individual agents and
Genetic Algorithm for reaching system equilibrium, minimizing the average travel time over different trips in the network. This method is thus able to achieve SE. \cite{mainali2008optimal} adopt Q-iterations with a model set-up similar to that of dynamic routing to be discussed next.

Most applications of RL to routing concern with the \emph{dynamic routing} (DR) problem (see Table \ref{tab:ref_routing}).  The MDP is modeled around a vehicle agent. The basic state information is the traffic state of the current node (i.e., intersection). Some works consider state features of the neighboring nodes \citep{kim2005optimal,mao2018reinforcement} so that the agent has a broader view of the environment. The action space comprises the set of outbound links (i.e., roads) or adjacent nodes from the current node, so the policy provides a turn-by-turn navigation guidance until the destination is reached. While it is most common to use travel time on a link as the reward function, \cite{tumer2008aligning,grunitzki2014individual} stand out by defining a new form called difference reward, which is the difference in average travel time on a link with and without the agent in the system. This applied to only a reward function dependent on the number of agents using the traversed link. In particular, travel distance cannot be used to define a difference reward. Whether solving a specific formulation achieves UE or SE depends on the reward function used. The average travel time on a link is a global reward because it is an aggregate of local rewards (i.e., individual travel times) of all the agents on that link. The difference reward, by definition, is a global reward that also reflects individual effect. If all the agents in the system learn by global reward \citep{tumer2008aligning,grunitzki2014individual,shou2020multi}, then the system is expected to achieve SE. Otherwise, the agents learn by their local rewards, and we will have UE or Nash equilibrium \citep{kim2005optimal,yu2012q,mao2018reinforcement,bazzan2016multiagent,wen2019hierarchical}.

Most works in the literature adopt Q-learning or its variant as the training algorithm. We report several notable developments. To tackle the sample efficiency issue of online model-free methods, \cite{mao2018reinforcement} propose an offline batch RL approach (fitted Q-iterations) with a tree-based function approximator (Extreme Randomized Trees) that empirically shows good convergence property. Hierarchical methods have also been adopted to address the complexity of a large-scale problem. In \citep{wen2019hierarchical}, the global road network is divided into sub-networks by differential evolution-based clustering.
The top-level network contains only the boundary nodes of the original network. The top-level policy produces the destination node for a sub-network. The sub-level policy provides link-level guidance to reach its sub-destination. \cite{shou2020reward} adopt a bilevel optimization scheme. At the lower level, a mean-field MARL algorithm solves for the dynamic routing problem for the travelers, while at the upper level, a Bayesian optimization module optimizes the control (i.e., reward parameter of the travelers) by the city planner.


\begin{sidewaystable}
\scriptsize
\begin{tabular}{||p{0.1\textwidth}|p{0.025\textwidth}|p{0.05\textwidth}|p{0.025\textwidth}|p{0.075\textwidth}|p{0.1\textwidth}|p{0.125\textwidth}|p{0.2\textwidth}|p{0.1\textwidth}||} 
\hline
Paper & Type & Equilibrium & Agent & State & Action & Reward & Algorithm & Road Network \\
\hline\hline
\cite{mainali2008optimal} & TAP & UE & vehicle & current node (intersection) & an adjacent link & travel time on the link & Q-iteration.
The route is constructed by following the decision at each intersection. & grid network \\
\hline
\cite{bazzan2015hybrid} & TAP & SE & vehicle & - & route from feasible routes for the OD pair & total travel time on the chosen route & Hybrid method:
Q learning for individual agent +
Genetic Algorithm for system equilibrium, minimizing avg travel time over different trips. & - \\
\hline
\cite{ramos2018analysing} & TAP & UE (empirical convergence) & vehicle & - & route from feasible routes for the OD pair & total travel time on the chosen route & Q-learning with action-regret updates to minimize driver's total regret & Braess graphs, OW network \\
\hline
\cite{zhou2020reinforcement} & TAP & UE (theoretical convergence established) & vehicle & - & route from feasible routes for the OD pair & total travel time on the chosen route & B-M RL scheme, similar to a MARL algo with individual reward. & Nguyen–Dupuis network \\
\hline
\cite{kim2005optimal} & DR & UE & vehicle & node, time, binary congestion status vector for all links & an adjacent node & cost accrued by traversing the link & parameters of MDP estimated from data, MDP solved by value iterations & Southeast Michigan network and traffic data \\
\hline
\cite{tumer2008aligning} & DR & SE & vehicle & just a single link & start time on the link & difference reward (uplift): with and without the agent in the system & Q-learning & - \\
\hline
\cite{yu2012q} & DR & UE & vehicle & current node & an adjacent link; action instruction received from the intersection in real time & travel time on the link & Similar approach to \citep{mainali2008optimal}, but incremental update is done by a step of SARSA.
The updates are done in real time and value functions are sync-ed at each intersection, which is an independent traffic management module. & SOUND/4U simulator based on the road network of Kurosaki, Kitakyushu in Japan \\
\hline
\cite{grunitzki2014individual} & DR & SE & vehicle & current node & an adjacent link & individual reward: negative travel time experienced by the agent on a link

difference reward: as in \citep{tumer2008aligning} & Applied to a more sophisticated network than aamas08 paper. Results show that DQ-learning outforms IQ-learning. & abstract network topology \\
\hline
\cite{mao2018reinforcement} & DR & UE & vehicle & current node, time, discrete congestion states vector of the arcs at most 2 steps away. & an adjacent node & negative travel time experienced by the agent on a link & offline batch RL: fitted Q-iterations with tree-based function approximator (Extreme Randomized Trees). Compared with model-based Q-iterations. & Sioux Falls network \\
\hline
\cite{bazzan2016multiagent} & DR & UE & trip (OD pair) & current node & an adjacent link & negative travel time experienced by the agent on a link & Independent Q learning
compared with successive average method & OW network, Sioux Falls network \\
\hline
\cite{wen2019hierarchical} & DR & UE & vehicle & next approaching node, destination node & an adjacent link & negative travel time experienced by the agent on a link & tabular Q-learning,
global road network clustered into subnetworks by differential evolution-based clustering

Top-level network contains only boundary nodes of the  original network. Top network policy produces the 'destination' node for a subnetwork. Subnetwork policy provides link-level guidance to reach its sub-destination. & SUMO simulator with various networks in Japan and US \\
\hline
\cite{shou2020reward} & DR & SE (avg travel time), UE (travel distance) & vehicle & current node, time & an adjacent link & negative average travel time of all agents or travel distance on a link & Bilevel optimization: 
Lower level - mean field MARL to solve for dynamic routing for travelers
Upper level - Bayesian optimization to optimize controls by city planners & SUMO with Manhattan network \\
\hline
\hline
\end{tabular}
\caption{Summary of literature for Route Guidance.}
\label{tab:ref_routing}
\end{sidewaystable}

\subsection{Ride-pooling (Carpool)}\label{sec:carpool}

\begin{figure}
\begin{center}
        \includegraphics[width=1.0\linewidth]{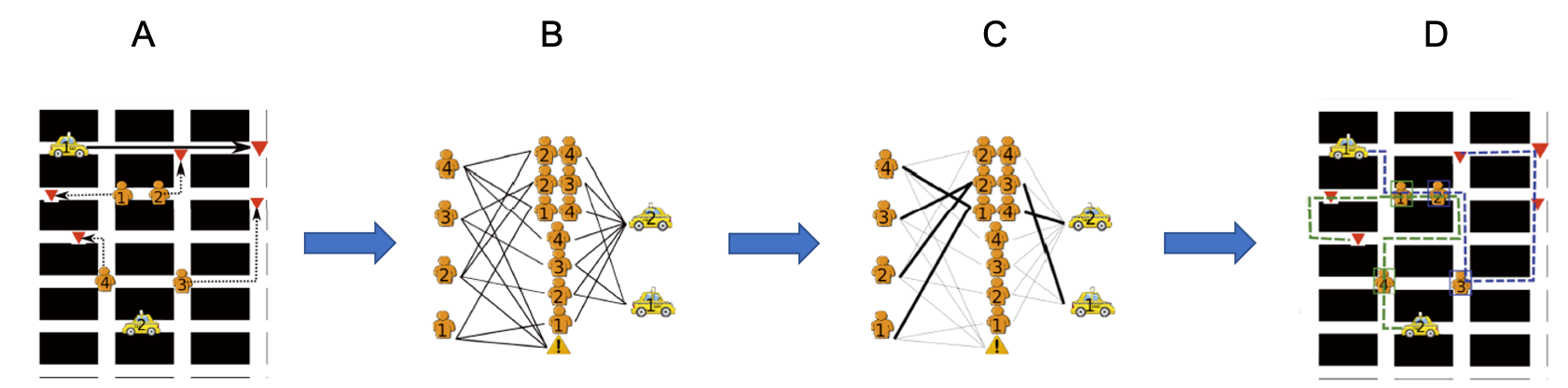}
	\caption{Illustration of the ride-pooling matching process adapted from \citep{alonso2017demand}. Stage A shows the state of the current vehicles and requests. Vehicle 1 has one passenger on board with her destination at the top right-hand corner, while vehicle 2 is empty. The feasible combinations of passengers and the vehicles feasible to serve them are determined at stage B. The corresponding assignment graph is set up and solved at stage C. Stage D shows the resulting routes to fulfill all four new requests as well as the existing trip.}
	\label{fig:pool_match}
\end{center}
\end{figure}

Ride-pooling optimization typically concerns with matching, repositioning, routing (see e.g., \citep{zheng2018order,alonso2017predictive,alonso2017demand,tong2018unified}). The RL literature  has primarily focused on the first two problems. The ride-pooling matching problem differs from that in Section \ref{sec:matching} in that a combination of multiple passengers, and hence their combined trip, can be matched to a vehicle that may or may not be empty. See stages B and C in Figure \ref{fig:pool_match} from \citep{alonso2017demand} for an illustration. The repositioning problem is similar to the ride-hailing case, except that the objective is to optimize some pooling-specific metrics that we define next. The routing problem solves for the sequence of pick-ups and drop-offs given the assigned passengers for a vehicle. The routing problem could also concern with route guidance on the road network. See stage D in Figure \ref{fig:pool_match}.

Many works have multiple objectives and define the reward as a weighted combination of several quantities, with hand-tuned weight parameters. \emph{Passenger wait time} is the duration between the request time and the pick-up time. \emph{Detour delay} is the extra time a passenger spends on the vehicle due to the participation in the ride-pooling. In some cases, these two quantities define the feasibility of a potential pooled trip instead of appearing in the reward \citep{shah2020neural}.
\emph{Effective trip distance} is the travel distance between the origin and destination of a trip request, should it be fulfilled without ride-pooling. \cite{yu2019integrated} consider minimizing passenger wait time, detour delay, and lost demand. 
\cite{gueriau2018samod} maximize the number of passengers served. \cite{jindal2018optimizing} maximize the total effective trip distance within an episode, which is just the number of served requests weighted by individual trip distance. Considering a fixed number of requests within an episode (hence fixed maximum effective distance), this metric reflects the efficiency of ride-pooling.  \cite{al2019deeppool,haliem2020dprs,singh2019distributed,haliem2020distributed} all attempt to minimize the SD mismatch, passenger wait time, reposition time, detour delay, and the number of vehicles used. 
In addition, \cite{haliem2020dprs} consider the fleet profit, and 
\cite{singh2019distributed} study a more general form of ride-pooling, where a passenger can hop among different vehicles to complete a trip, with each vehicle completing one leg. They further consider the number of hops and the delay due to hopping.

The state of an agent usually consists of global SD information, similar to that for matching and reposition, but the vehicle status contains key additional information of occupancy and OD's of the passengers on board.

The action space depends on whether the agent is modeled at vehicle level or system level. Existing RL-based works all require that a vehicle drops off all the passengers on board according to a planned route before a new round of pooling. An individual vehicle agent can then match to a feasible group of passengers (in terms of capacity and detour delay) \citep{jindal2018optimizing}, reposition to another location \citep{al2019deeppool,haliem2020dprs,haliem2020distributed}, or both \citep{gueriau2018samod}. A system-level agent has to make action decisions for the entire fleet together \citep{yu2019integrated,shah2020neural}. The feasible combinations of passengers are typically determined by a separate process based on a pairwise shareability graph or a trip-vehicle graph\citep{alonso2017demand}. (See illustration in Figure \ref{fig:pool_match}.)

Papers with vehicle-level policy commonly train a single agent and apply to all the vehicles independently (see e.g., \citep{haliem2020distributed}).
DQN is a convenient choice of training algorithm for this setting. For system-level decision-making, both \cite{yu2019integrated} and \cite{shah2020neural} employ an ADP approach and consider matching decisions only. \cite{yu2019integrated} follow a similar strategy as \citep{simao2009approximate} and use a linear approximation for the value function. In contrast, \cite{shah2020neural} decompose the system value function into vehicle-level ones and adopts a neural network for the individual value function, which is updated by mini-batch stochastic gradient descent similar to that in DQN. 

It has become increasingly clear that dynamic routing and route planning in the context of ride-pooling require specific attention. In particular, there are two aspects unique to ride-pooling. First, the trips are known only at their request times. Hence, the routes taken by the pooled vehicles (i.e., the sequences of pick-ups and drop-offs) have to be updated dynamically to account for the newly joined passengers. \cite{tong2018unified, xu2020efficient} formulate the route planning problem for ride-pooling and develop efficient DP-based route insertion algorithms for carpool. In Section \ref{sec:vrp}, we will see that this is also part of the stochastic dynamic vehicle routing problem. Second, within a given route plan, the route taken by a pooled vehicle from an origin to a destination can affect the chance and quality of its future pooling. Hence, dynamic routing (or route choice) between an OD pair can be optimized in that direction, e.g.,  \cite{yuen2019beyond} go beyond the shortest-path to make route recommendations for better chance of pooling. \cite{gueriau2020shared} evaluate the SAMoD system proposed in \citep{gueriau2018samod} in a microscopic environment based on SUMO with a traffic congestion-aware (non-RL) routing component. We expect to see more RL-based algorithms for the ride-pooling dynamic routing problems.

\subsection{Vehicle Routing Problem (VRP)}\label{sec:vrp}
VRP has close connection to the ridesharing problem in that variants of VRP could serve as a subroutine of the ride-pooling problem and could even used to model the entire ridesharing problem itself. The main challenge, in the context of ridesharing, is that new demand (a pair of pick-up and drop-off locations) appears in an online nature and has to be inserted into the existing route dynamically. So reviewing the RL literature for VRP is not only for the completeness of this survey but also essential for one to appreciate the complexity and challenges in tackling ridesharing via RL.

VRP has many variants in its rich literature, so it is important to be clear on their differences and on the variant that each paper claims to solve. The basic setup of a VRP consists of a transportation network $\mathcal{G}:=(\mathcal{V},\mathcal{E},w)$ and a fleet of $K$ vehicles, where $\mathcal{V}$ is the set of nodes (customer and depot locations), and $\mathcal{E}$ is the set of edges such that $e_{ij} \in \mathcal{E}$ indicates that it is possible to travel from node $v_i$ to node $v_j$. $w(x_{ij}):=w_{ij}$ is the edge cost, typically distance or travel time. The depot $v_0 \in \mathcal{V}$ is a special node where all the vehicles depart and return at. The vanilla VRP is to find an optimal set of disjoint routes (one for each vehicle) that start and end at the depot, that collectively cover all the nodes, and whose total cost (summing over all the edges in those routes) is minimized. A \emph{traveling salesman problem} (TSP) is a special (simplified) instance of VRP, in which there is no depot, and the fleet consists of a single vehicle. In a \emph{capacitated} VRP (CVRP), each node has a demand quantity to be fulfilled by one of the vehicles. Each vehicle has a limited capacity, and it starts from the depot with a full load of goods to fulfill the demand of the nodes on its route. The total capacities of the fleet is sufficiently large, and all the demand has to be fulfilled. A CVRP with \emph{split delivery} allows the demand of each node to be fulfilled by multiple vehicles. In a VRP with \emph{time windows}, each node has a delivery window within which the node has to be visited or its demand has to be satisfied. If the time window constraints are \emph{soft}, they can be violated with the price of a penalty that contributes to the total cost function. In some variants of the VRP, the goods for the demand of a node (destination) has to be picked up from another designated non-depot location (origin) before being delivered to it. This is known as a VRP with \emph{pick-up and delivery}, also known as the \emph{dial-a-ride} problem (DARP). If the fleet consists of electric vehicles (EV), the set of nodes also include charging stations, and each EV has a limited battery capacity, before the depletion of which the EV has to reach a charging station to recharge. In practical situations, a VRP or variant can be \emph{stochastic} and \emph{dynamic} (SDVRP), i.e., its parameters (e.g., demand and travel time) are uncertain, and the requests are not known at the beginning but are revealed sequentially throughout the problem period. 

The connection between VRP and ridesharing exists at both local and fundamental levels. As a subproblem in ride-pooling, the rerouting problem after a new passenger is matched to the vehicle is a TSP with pick-up and delivery (TSPPD), which is one-vehicle single-tour instance of CVRP with pick-up and delivery.  At a fundamental level, (multi-vehicle) ride-pooling is a stochastic dynamic multi-vehicle CVRP with pick-up and delivery. Although there are no explicit time windows, cancellation may occur if waiting time is too long. Ride-hailing is also a special case where the vehicles all have unit capacity, and in this case, matching and routing merge into one single problem. So the ridesharing problem is an SDCVRP, except that repositioning is an intervention strategy not considered in SDCVRP.

The goal of this section is not to provide a complete survey of the VRP literature but rather to point out the representative or unique works that adopt RL to solve VRPs (see Table \ref{tab:ref_vrp}).
A recent review of RL-based methods for solving stochastic dynamic VRP can be found in [Hildebrandt, et al., 2021] and a more general one in \citep{ulmer2020modeling}.

\paragraph{Single-vehicle v.s. multi-vehicle problems}
CVRP may appear in different forms, and sometimes the subtle differences may not be stated clearly.
Most papers solve the single-vehicle problem where there is only one active vehicle at any time. In the capacitated single vehicle problem, the vehicle can make multiple tours (i.e., passing through the depot multiple times) to fulfill all the demand, but the number of tours is not set in advance. In the multi-vehicle case, a fleet of $K$ vehicles are active simultaneously. For static VRPs, if the number of tours in the single-vehicle case is fixed, then it is equivalent to the multi-vehicle counterpart by treating each tour as a separate vehicle. (For problems with time windows, this equivalence can be achieved by resetting the clock every time a new tour starts.) Otherwise, they are not equivalent in general because the number of tours in the optimal solution for the single-vehicle problem may not be $N$. For dynamic problems where the requests are not all known a-priori, it is not possible to generate the fixed number of tours in sequence, since one cannot insert a new request to a previous tour. In this case, equivalence can only be achieved by keeping each tour on the same clock and updating the routes with the newly appeared requests at each time step. As we will see below, this would render essentially a multi-vehicle algorithm.

The majority of the RL-based methods for VRP models the agent as a vehicle with the system-state visibility.  The state thus consists of two types of information: the vehicle state, which includes the vehicle's current location and remaining capacity (for pick-up and delivery, e.g., \citep{ulmer2020modeling,james2019online,joe2020deep}) or inventory (for homogeneous goods delivery, e.g., \citep{nazari2018reinforcement,kool2018attention,delarue2020reinforcement}); the system state, which contains the locations of the customer nodes, the demand at each node, and the unserved customers. For pick-up and delivery problems, the system state instead contains the pick-up and delivery locations of the orders. In the case of EVs, the vehicle state additionally contains the vehicle's battery level, and the system state also includes the locations of the charging stations and the number of vehicles available (not in charging). The action of the agent is to specify the next stop (pick-up/delivery location or charging station in the case EV) to visit for the current vehicle. The sequence of actions form a route for the vehicle. When multiple routes/tours are required, the different routes are separated by the insertion of the depot \citep{nazari2018reinforcement,duan2020efficiently,kool2018attention,lin2021deep}. For (dynamic) multi-vehicle problem where decisions for all the vehicles are made at each time step, the agents would generate their actions sequentially to avoid conflicting actions \citep{james2019online,zhang2020multi}.
Since the objective of VRPs is typically to minimize total travel distance, the reward is naturally defined as the negative travel distance. For problems with (soft) time window constraints, the negative penalty for constraint violation is added to the reward.

Typically, these methods adopt an encoder-decoder agent network architecture. The encoder is responsible for encoding part or all of the state information into an embedding vector (or context), which, potentially with additional input state features, is fed into the decoder to generate the action one at a time. 
\cite{bello2016neural} develop an policy network based on the pointer network \citep{vinyals2015pointer}, which consists of an RNN encoder and an RNN decoder. The major novelty over a sequence-to-sequence architecture is that the decoder uses attention mechanism to attend over the embeddings of the input nodes to generate the probability distribution over the input space, thus eliminating the distance disparity in the output with respect to the input, a feature that is particularly suitable for solving TSP and VRP. To reduce the complexity of the encoder and avoid imposing a sequence on the input state features (e.g., customer locations), which is unnecessary in routing problems, \cite{nazari2018reinforcement} modify the pointer network with a non-sequential encoder which simply embeds each individual input node. They incorporate the policy network into an AC method and validate the design on a CVRP with split delivery. A few more recent works have adopted this network structure. \cite{james2019online} use structural graph embedding (Struct2Vec) for the encoder, since their agent's state additionally contains a vehicle tour graph. 
In \citep{lin2021deep}, the encoder has 1D convolution and graph embedding for the input nodes, followed by an attention layer. \cite{duan2020efficiently} include edge features in the state besides the node features of the transportation network. Their encoder is based on graph convolution network with both node and edge inputs.  Another work with significant novelty is \citep{kool2018attention}, which develops a policy network with a transformer-based encoder and a self-attention-based decoder to use in a PG method (REINFORCE) with the baseline computed from deterministic greedy rollout. This training framework has also been adopted by \cite{zhang2020multi} for multi-vehicle VRP with soft time windows, \cite{lin2021deep} for EV VRP with time windows, and \cite{duan2020efficiently}, which jointly train an MLP-based binary classifier on edge encoding with the policy network output as labels. They have tested their method on a CVRP with 400 nodes, the largest among the reviewed works.

For SDVRP, \cite{ulmer2020modeling} argue that it is a more convenient model, which also aligns better with popular approaches to this problem, that the action contains also the route plan information. They define a new variant of MDP, called route-based MDP, in which the state includes the route plan from the last epoch, and the action contains the updated route plan in addition to the next stop to visit. The `immediate' reward becomes the difference in route value between the old and new plans. Following this line, \cite{joe2020deep} model a system agent whose state includes the cost for the remaining route for each vehicle, and the agent assigns a new request to a vehicle at each decision epoch. The rerouting after matching is solved by simulated annealing for VRP. Under this framework, one only needs to learn an action-value function to generate the matching decisions. The algorithm is tested on a multi-vehicle SDVRP with pick-up/delivery and time windows.\footnote{This method can be regarded as one for the matching problem in ride-pooling described in Section \ref{sec:carpool}.} In a somewhat similar spirit but for static CVRP, the MDP action in \citep{delarue2020reinforcement} is to generate one route (tour). The value network consists of dense layers and ReLU activation and is representable by mixed-integer linear constraints so that the action can be computed through solving a Prize Collecting TSP by MIP.

\begin{sidewaystable}
\scriptsize
\begin{tabular}{||p{0.1\textwidth}|p{0.08\textwidth}|p{0.1\textwidth}|p{0.1\textwidth}|p{0.1\textwidth}|p{0.2\textwidth}|p{0.1\textwidth}|p{0.1\textwidth}||} 
\hline
Paper & Type & State & Action & Reward & Network & Algorithm & Problem Size \\
\hline\hline
\cite{nazari2018reinforcement} & single-vehicle & location and demand of each request & the next request to visit & negative travel distance & Non-sequential encoder for the input with RNN decoder that attends over over the input space (Pointer network without an RNN encoder) & AC & Single-vehicle Capacitated VRP with split delivery: one active vehicle at a time \\
\hline
\cite{kool2018attention} & single-vehicle & coordinates and original demand of each node, remaining demand of each node, remaining capacity of the vehicle & next stop for a given vehicle & negative travel distance & transformer encoder with input of coordinates and demand of each node + self-attention-based decoder with additional input of remaining demands and capacity at time t & REINFORCE with greedy rollout baseline & TSP and Capacitated VRP with split delivery, 100 nodes \\
\hline
\cite{balaji2019orl} & single-vehicle  & current pickup location, vehicle's location and remaining capacity, orders' locations, statuses, waiting times, and values & accept an order, pick up an accepted order, wait & value of delivered order (accept, pickup, deliver) - cost (waiting, traveling, penalty) & two dense layers NN & APE-X DQN \citep{horgan2018distributed}, a variant of a DQN that utilizes distributed prioritized experience replay  & stochastic and dynamic CVRP with pick-up/delivery and time windows, 8 x 8 map, 5 orders 3 pick-up locations \\
\hline
\cite{james2019online} & multi-vehicle & system state (available requests, charging station output, vehicles' status(location, battery levels, next stops)),  vehicle tour graph & next stop for a given vehicle; The vehicles generate actions sequentially at each time step. & expected objective value for a tour & pointer network for actor, another critic network; 
Network architecture is similar to NeurIPS-18 paper, but with structural graph embedding (Struct2Vec) for the encoder. & A3C & multi-vehicle dynamic VRP with pick-up and delivery for EVs: 200 random requests, 100 vehicles \\
\hline
\cite{zhang2020multi} & multi-vehicle & same as \citep{kool2018attention} & same as \citep{kool2018attention}; The vehicles generate actions sequentially at each time step. & negative travel distance + negative constraint violation penalty & same as \citep{kool2018attention} & same as \citep{kool2018attention} & Multi-vehicle VRP with soft time windows (no split delivery): 150 nodes, 5 vehicles \\
\hline
\cite{ulmer2020modeling} & single-vehicle & vehicle location, time, num of passengers onboard, info of in-process and outstanding requests, route plan from last epoch & the next stop to visit and the new route plan; The paper defines a new variant of MDP called route-based MDP. & difference in route value between the old route plan and the new one & N.A. & insert new request s into the current route and use variable neighborhood search to improve the route & SDVRP with pick-up and delivery (DDARP) \\
\hline
\cite{joe2020deep} & multi-vehicle & includes the cost for the remaining route for each vehicle & matching a new order to a vehicle. Rerouting after matching is done by simulated annealing for VRP & cost diff between two consecutive decisions & not specified & NN-based TD learning with experience replay (like in \citep{tang2019deep}) to learn the action-value function & Multi-vehicle dynamic VRP with pick-up/delivery and delivery windows: 48 nodes, 2 vehicles, avg 22 orders/day \\
\hline
\cite{delarue2020reinforcement} & single-vehicle & the remaining unvisited nodes & to generate one route (tour) through solving a Prize Collecting TSP by MIP & negative tour distance & Value network consists of dense layers + ReLU activation (representable by mixed-integer linear constraints) & MC policy iteration: rollout N trajectories, fit a new NN & CVRP: 51 nodes \\
\hline
\cite{duan2020efficiently} & single-vehicle & nodes (location, demand), edges (distance, adjacency) & generate one node at a time sequentially; The resulting sequence may have multiple depot occurrences for different tours. & negative travel distance & GCN-based encoder with both node and edge features; GRU-based decoder similar to the pointer network as policy network and MLP-based decoder on the edge encoding as classifier & REINFORCE with greedy rollout baseline \citep{kool2018attention} to train the policy network; Cross-entropy loss to train the binary classifier of route edges with policy output as labels & CVRP: 400 nodes \\
\hline
\cite{lin2021deep} & multi-vehicle & For time $t$, the state of each vertex (location, time window, remaining demand), and global variables (time, battery level of the active vehicle, number of EVs not in charging) & Next stop for the current route; 
Unlike \citep{james2019online} the routes of the vehicles are generated sequentially. Every time the depot appears in the sequence, the system time is reset to 0. & negative travel distance + negative penalties for constraint violations & Encoder with 1D conv layer and graph embedding for the nodes and attention layer; LSTM-based decoder & REINFORCE with greedy rollout baseline \citep{kool2018attention} & EV with time window and charging.
Within the planning horizon, a vehicle can visit the depot only once: C100, S12, EV12 \\
\hline
\hline
\end{tabular}
\caption{Summary of literature for VRP.}
\label{tab:ref_vrp}
\end{sidewaystable}

\subsection{Data Sets \& Environments}\label{sec:data}
The problems in ridesharing are highly practice-oriented, and results from toy data sets or environments may present a very different picture from those in reality. Hence, real-world data sets and realistic simulators backed up by them are instrumental to research in RL algorithms for these problems. 

The most commonly used data sets are those made available by NYC TLC (Taxi \& Limousine Commission) \citep{nycdataset}. This large public data repository contains trip records from several different services, Yellow Taxi, Green Taxi, and FHV (For-Hire Vehicle), from 2009 to 2020. The Yellow Taxi data is the most frequently used for various studies. The FHV trip records are submissions from the TLC-licensed bases (e.g., Uber, Lyft) and have a flag indicating pooled trips offered by Uber Pool and Lyft Line. 
The pick-up and drop-off locations are represented by taxi zones. 
Manhattan, for example, is divided into 64 zones. 
There is no driver ID associated with the trip records, so reconstructing historical driver-based trajectories is not possible. 
An older version of the NYC data set \citep{illinoisdatabankIDB-9610843}, however, does include GPS coordinates for pick-up and drop-off locations, and car IDs can be used to track drivers within each year, allowing for more granular and diverse analyses.
A similar subset of the NYC FHV data is also available at \citep{fivethirtyeight}, with GPS coordinates for pick-up and drop-off locations. In addition, travel time data between OD pairs can be obtained through Uber Movement \citep{ubermovement}. 

Another taxi data set is the San Francisco data set \citep{epfl-mobility-20090224}, which contains GPS coordinates of approximately 500 taxis collected over 30 days in the San Francisco Bay Area in May 2008. The average time interval between two consecutive location updates is less than 10s. 

A more recent rideshare (Transportation Network Providers, TNPs) data set is published by Chicago Data Portal \citep{chicago2018data}. This data set contains trips, drivers, and vehicles data reported by Transportation Network Providers (TNP, or rideshare companies) in Chicago from 2018. Trip origin and destinations are represented by census tracts. Times are rounded to the nearest 15 minutes. Fares are rounded to the nearest \$2.50 and tips are rounded to the nearest \$1.00. The driver and vehicle data are not joinable with the trip data. 

Developing ridesharing simulators has been a line of research itself. \cite{yao2021ridesharing} offer a comprehensive review of recent works on ridesharing simulation models, most of them covering a subset of considerations on the number of passengers, the pre-/post-match passenger cancellation behaviors, and driver acceptance/rejection behaviors. In \citep{yao2021ridesharing}, a sophisticated event-based simulation framework is proposed to capture all aspects of the behavior modeling. Although the `ridesharing' in their paper is known as the hitch service, where the driver is on her own trip as well, the modeling framework is general and accommodates the ridesharing setting in this survey. 
\cite{chaudhari2020learn} offer a Gym-compatible, open-source ride-hailing environment \citep{learn2earn2020code} for training dispatching and repositioning agents. For large-scale simulation on transport networks, AMoDeus \citep{ruch2018amodeus} and MATSim \citep{w2016multi} are well-established Java-based simulation frameworks that also come with graphical user interfaces and visualization tools. They are of more sophisticated engineering architectures albeit with higher programming bars for extension.
The evaluation simulation environment for the KDD Cup 2020 competition is available for public access through the DiDi decision intelligence simulation platform \citep{didisim}. Although not yet open-sourced, this simulation environment supports both matching and vehicle repositioning tasks and accepts input algorithms through a Python API.

\section{Challenges and Opportunities}\label{sec:discussions}
Given the state of the current literature, we discuss a few challenges and opportunities that we feel crucial in advancing RL for ridesharing.


\subsection{Ride-pooling} 

As seen in Section \ref{sec:carpool}, the reward function in ride-pooling is often a hand-tuned combination of multiple objectives. It is desirable to have a principled way to determine the best weighting scheme automatically, potentially leveraging inverse RL and multi-objective learning techniques \citep{zou2021dynamic,arora2021survey} in a similar sense of the ride-hailing case \citep{zhou2021multi}.
Methods for learning to make matching decisions are still computationally intensive \citep{shah2020neural,yu2019integrated}, in part due to the need to use VRP solver to determine feasible actions (combination of passengers). Moreover, all existing works assume that the action set is pre-determined, and some make only high-level decisions of reposition and serving new passengers or not. A more sophisticated agent may be called for to figure out, for example, how to dynamically determine the desirable passenger combination to match to a vehicle and the routes to take thereafter. Ride-pooling pricing \citep{ke2020pricing}, a hard pricing problem itself, is tightly coupled with matching. A joint pricing-matching algorithm for ride-pooling is therefore highly pertinent. As mentioned in Section \ref{sec:carpool}, it is also highly anticipated to go beyond using generic routing algorithms and to tailor them to ride-pooling with RL.

\subsection{Joint Optimization}
The rideshare platform is an integrated system, so joint optimization of multiple decision modules leads to better solutions that otherwise unable to realize under separate optimizations, ensuring that different decisions work towards the same goal. RL for joint optimization across multiple modules calls for research on reward function design, state-action representation that facilitates inter-module communication, and the training algorithms. Models and algorithms that allow decentralized execution by the different modules are highly preferred in practice.
We have already seen development on RL for joint matching-reposition \citep{holler2019deep,jin2019coride,tang2021value} and with ride-pooling \citep{gueriau2018samod}, pricing-matching \citep{chen2019inbede}, and pricing-reposition \citep{turan2020dynamic}. An RL-based method for fully joint optimization of all major modules is highly expected. Meanwhile, this also requires readiness from the rideshare platforms in terms of system architecture and organizational structure.

\subsection{Heterogeneous Fleet}
With the wide adoption of electric vehicles and the emergence of autonomous vehicles, we are facing an increasingly heterogeneous fleet on rideshare platforms.
Electric vehicles have limited operational range per their battery capacities. They have to be routed to a charging station when the battery level is low (but sufficiently high to be able to travel to the station). Autonomous vehicles may run within a predefined service geo-fence due to their limited ability (compared to human drivers) to handle complex road situations. For an RL-based approach, a heterogeneous fleet means multiple types of agents with different state and action spaces. The adoption of autonomous vehicles also opens new operational paradigms. Dynamic fleet size inflation \citep{beirigo2022business}, for example, hires idle autonomous vehicles on demand to guarantee service quality contracts in a ridesharing marketplace.
Specific studies are required to investigate how to make such a heterogeneous fleet cooperate well to complement each other and maximize the advantage of each type of vehicles to improve overall system efficiency.

\subsection{Simulation \& Sim2Real}
Simulation environments are fundamental infrastructure for successful development of RL methods. Despite those introduced in Section \ref{sec:data}, simulation continues to be a significant engineering and research challenge. We have rarely seen comparable simulation granularity as that of the environments for traffic management, (e.g., SUMO \citep{SUMO2018}, Flow \citep{wu2017flow}) or autonomous driving (e.g., SMARTS \citep{zhou2020smarts}, CARLA \citep{dosovitskiy2017carla}).\footnote{\cite{gueriau2020shared} evaluate the ridesharing algorithms in SUMO, but the environment is not public.} 
The opportunity is an agent-based microscopic simulation environment for ridesharing that accounts for both ride-hailing and carpool, as well as driver and passenger behavior details, e.g., price sensitivity, cancellation behavior, driver entrance/exit behavior. None of the existing public/open-source simulators supports pricing decisions. Those simulators described in the pricing papers all have strong assumptions on passenger and driver price elasticities.
A better way might be to learn those behaviors from data through, e.g., generative adversarial imitation learning \citep{shang2019environment} or inverse RL \citep{mazumdar2017gradient}.

No publicly known ridesharing simulation environment has sufficiently high fidelity to the real world to allow an agent trained entirely in it to deploy directly to production. Several deployed works \citep{qin2020ride,jtq2021repos} in Section \ref{sec:survey} have all adopted offline RL for learning the state value functions and online planning. The robotics community has been extensively investigating ways to close the reality gap \citep{traore2019continual,mehta2020curriculum}. Sim2real transfer algorithms for ridesharing agents are urgently sought after.



\subsection{Human Behavior}\label{sec:behavior}
Central to ridesharing platforms are human participants (passengers and drivers).\footnote{A partial exception is an autonomous ridesharing platform, where the supply side is powered by autonomous vehicles. However, such services are still prototypical at the time of writing and have very limited coverage.} 
The impact of human behavior is pervasive in the ridesharing marketplace, e.g., in request conversion, cancellation, idle driver diffusion, driver sign-in and sign-off, rider and driver responses to incentives. Human behavior is inherently stochastic and difficult to model, especially with limited data (in size and features), which introduces errors to optimization and simulation. Compared to traditional approaches from operations research, RL offers potential to better handle these stochasticity issues through its adaptability and data-driven nature.

Unlike cumulative effects induced by spatiotemporal transitions (e.g., matching), human-induced long-term effects from changes in habituation and sentiment on the marketplace are much harder to learn due to the much longer horizon such effects span over. To RL, this is dual challenge and opportunity. The challenge is the long feedback loop and very delayed reward signals, and the opportunities lie in engineering and capturing more refined system state features that capture human behavior characterization better and in designing a richer set of reward signals that facilitate the learning of policies for long-term optimality.

\subsection{Non-stationarity}
We have seen in Sections \ref{sec:matching} and \ref{sec:reposition} that RL algorithms deployed to real-world systems generally adopt offline training - once the value function or the policy is deployed, it is not updated until the next deployment. Value functions trained offline using a large amount of historical data are only able to capture recurring patterns resulted from day-on-day SD changes. However, the SD dynamics can be highly non-stationary in that one-time abrupt changes can easily occur due to various events and incidents, e.g., concerts, matches, and even road blocks by traffic accidents. To fully unleash the power of RL, practical mechanisms for real-time on-policy updates of the value function (e.g., \citep{tang2021value,tong2021combinatorial,eshkevari2022reinforcement}) is required. In view of the low risk tolerance of production systems in general, sample complexity, computational complexity, and robustness are the key challenges that such methods have to address.

\subsection{Business Strategies}
The research problems in the ridesharing domain are closely associated with how the ridesharing platforms run the operations. Innovation in product and business operations will continue to raise new challenging research problems. There can be multiple alternative product forms to achieve the same goals or address the same challenges, and they inherently define different optimization problems that RL can help tackle. 

Surge pricing, for example, is a pricing strategy during the peak hours to address the severe shortage of supply with respect to the surging demand. We have explained its motivation in Section \ref{sec:pricing}. While surge pricing is a common practice nowadays, it is not the only strategy that the ridesharing platforms adopt. Passenger requests can be queued if there are no vacant vehicles around to immediately serve the requests \citep{zhong2020queueing}. 
The queuing mechanism is perceived in some markets as a more socially acceptable mechanism during the peak hours than surge pricing. Several operational decision questions immediately come up, e.g., how large an area each queue should cover, if the coverage should be dynamically updated, and when the incoming requests should start queuing. These potentially time-varying decisions in a highly stochastic environment are good candidates to be solved for by RL.

Ridesharing platforms often use incentives to stimulate growth on both sides of the marketplace. The forms of incentives are diverse and ever evolving: rider coupons, discounts, target-based challenges, driver bonuses with spatial and temporal constraints, etc. Each incentive strategy changes the behaviors of a certain segment of the marketplace participants in a certain way, and they inevitably interact with the other marketplace levers, e.g., dynamic pricing \citep{yang2020integrated}. The collective effects of the evolving incentives convolute the environment and dynamics of the marketplace, posing significant challenges to RL and other optimization methods. How to represent and capture these factors or explicitly model them in joint optimization is key to tackle these challenges.

Third-party service integrator allows passengers to simultaneously request orders from multiple ride-hailing platforms \citep{zhou2020competitive}. Service integrators offer the platforms more access to the demand but also bring competition more explicit by displaying the matching information (e.g., trip fare, pick-up distance) side by side. Optimizing pricing and matching policies in a competitive environment with feedback from the service integrator on the competition landscape will be interestingly different from those without a service integrator or in a non-competitive environment. With the added environment complexity, these problems are challenging to solve by traditional methods and could be better tackled by RL. 

\subsection{General RL}
RL provides the necessary tools for the methods reviewed in this survey. Hence, the problems of RL for ridesharing tie closely to the development in RL in general. In the context of ridesharing, we have seen from the literature review above that it is difficult for RL to learn combinatorial actions, e.g., the system matching actions. In the era of deep RL, model interpretability is a long-standing challenge, which hampers investigation of customer experience corner cases. For experience-critical service like ridesharing, policy exploration adds further complication, especially for real-world deployment.
In view of these challenges, the future is probably that RL-based and traditional optimization approaches will be complementing each other for a long time. We have seen such combinations in the current literature as \citep{xu2018large,qin2021reinforcement} for matching, \citep{chaudhari2020learn,jtq2021repos} for repositioning, and \citep{delarue2020reinforcement} for VRP, that combine RL with combinatorial optimization, mixed-integer programming, and tree search. 
The breakthroughs of RL that we are seeing in other domains and the continued development of RL methodology for ridesharing certainly make it exciting to anticipate the future landscape.

\section{Closing Remarks}
We have surveyed the RL literature for the core problems in ridesharing: pricing, dispatching, repositioning, routing, ride-pooling, and VRP. We have also discussed some open challenges and future opportunities pertinent to this area. 

The ridesharing system is a complex multi-agent system with multiple decision levers. RL offers a powerful modeling vehicle for optimizing this system, but as we have seen from the current literature, challenges remain in tackling complexity in the learning algorithms, the coordination among the agents, and the joint optimization of multiple levers. Along tackling these challenges, we expect that domain knowledge in ridesharing as well as transportation in general will be increasingly instrumental to the successful adoption of RL. As one may have noticed, most of the literature has just appeared in the last four years, and we expect it to continue growing and updating rapidly. 

\bibliographystyle{agsm}
\bibliography{tony_bib}

\end{document}